\documentclass[11pt,a4paper]{article}

\usepackage[comma]{natbib}
\usepackage{anysize}
\usepackage{color}
\usepackage{graphicx}
\usepackage{pdfsync}

\usepackage{amsmath,amssymb,amsthm,amsopn}
\usepackage[ruled]{algorithm2e}

\theoremstyle{remark}

\renewcommand{\today}{\begingroup
\number \day\space  \ifcase \month \or January\or February\or
March\or April\or May\or June\or July\or August\or September\or
October\or November\or December\fi \space  \number \year \endgroup}

\theoremstyle{plain}

\newtheorem{teor*}{Teorema}

\theoremstyle{definition}

\title{Mixture model modal clustering}

\author{Jos\'e E. Chac\'on\footnote{Departamento de
Matem\'aticas, Universidad de Extremadura, E-06006 Badajoz, Spain. E-mail:
{\tt jechacon@unex.es}}}

\begin{document}

\maketitle

\begin{abstract}
\noindent The two most extended density-based approaches to clustering are surely mixture model clustering and modal clustering. In the mixture model approach, the density is represented as a mixture and clusters are associated to the different mixture components. In modal clustering, clusters are
understood as regions of high density separated from each other by zones of lower density, so that they are closely related to certain regions around the density modes. If the true density is indeed in the assumed class of mixture densities, then mixture model clustering allows to scrutinize more subtle situations than modal clustering. However, when mixture modeling is used in a nonparametric way, taking advantage of the denseness of the sieve of mixture densities to approximate any density, then the correspondence between clusters and mixture components may become questionable. In this paper we introduce two methods to adopt a modal clustering point of view after a mixture model fit. Numerous examples are provided to illustrate that mixture modeling can also be used for clustering in a nonparametric sense, as long as clusters are understood as the domains of attraction of the density modes.
\end{abstract}


\newpage

\section{Introduction}

Classical clustering algorithms are mainly based on inter-point distances (e.g., hierarchical clustering) or on partitioning the space around a pre-fixed number of central points (these are usually called partitioning methods, and include $K$-means clustering, for instance). In the recent times, however, there is a growing body of researchers that advocate that ``density needs to be incorporated in the clustering procedures" \citep{CM13}.

Two very different density-based approaches to clustering are mixture model clustering \citep{FR02} and clustering based on high density regions \citep{H75}. The former, in a parametric context, starts by modeling the distribution density $f$ as a mixture of densities in a pre-specified parametric family, that is, $f(x)=\sum_{g=1}^G\pi_gf_g(x)$ where the mixing weights $\pi_g>0$ are such that $\sum_{g=1}^G\pi_g=1$ and the density components $f_1,\dots,f_G$ can be written as $f_0(x|\theta_1),\dots,f_0(x|\theta_G)$ for a fixed parametric distribution $f_0(x|\theta)$ and different parameter values $\theta_1,\dots,\theta_G$. If this mixture model is identifiable, it seems natural to associate different clusters to each of the distribution components. In practice, a density estimate $\widehat f$ within this model is obtained by estimating the parameters and mixing weights by maximum likelihood, and selecting the number of components using the Bayesian information criterion (BIC) \citep[again, details can be found in ][]{FR02}, leading to $\widehat f(x)=\sum_{g=1}^{\widehat G}\widehat \pi_gf_0(x|\widehat\theta_g)$. Then, through Bayes theorem, any point $x$ in the space can be assigned to the component that makes it more probable by looking at the value of $g\in\{1,\dots,\widehat G\}$ that maximizes $\widehat \pi_gf_0(x|\widehat\theta_g)$.

The principles of clustering based on high density regions are quite different. In this context, clusters are understood as regions of tight concentration of probability mass, separated by each other by regions where the probability mass is more dispersed. There are two ways to formalize this. \citet{H75} proposed to focus on the region where the density is above some pre-specified level (density level sets) and defined clusters as the connected components of this region. This clearly captures the notion of a high density region (the density must be above some level) separated by regions of lower density (this happens where the region consists of more than one connected component). The main disadvantages of this definition are: first, since it concerns the region where the density is above some level, it may leave a substantial number of points with no cluster assigned; second, the whole cluster structure of the distribution may not be noticeable at a fixed, single level \citep[see, for instance, Figure 1 in][]{RSNW12}; and third, the obtention of the connected components of a density level set is not a computationally easy task (see \citealp{CFF01}, or \citealp{AT07}). The first and second issues can be amended by considering the cluster tree \citep{S03}, which represents how the density level clusters evolve as the level varies (in a close connection to persistent homology techniques, see \citealp{EH08}). This solution, unfortunately, emphasizes the aforementioned computational issue, since the construction of the cluster tree involves computing the connected components of not just one, but several density level sets.

An alternative formalization of high density clusters is through Morse theory tools. Clusters are defined as the domains of attraction of the density modes; i.e., a cluster is made of all the points that are eventually taken to a given local maximum of the density, when moved through the flow line defined by the density gradient field \citep{Ch15}. Having the clusters closely connected to the density modes, this approach is commonly known as modal clustering. It is related to, but different from the notion based on the connected components of density level sets, and avoids the above noted drawbacks of the latter. This definition results in a partition of the whole space into clusters, provided the density is sufficiently regular, with the boundaries of the partition components made of density valleys (regions of lower density), it does not require the choice of a level parameter and it is computationally tractable though the adaptation of numerical optimization methods to this setting, such as the mean shift algorithm \citep{FH75}, a variant of the gradient ascent algorithm for function maximization.

Any of the two high density clustering approaches is described above in population terms, that is, in terms of the true density $f$. If a $d$-variate sample $X_1,\dots,X_n$ from $f$ is given, then empirical, data-based clusters are obtained by replacing the unknown underlying density by a density estimate $\widehat f$. Since no parametric model is assumed in this setting, it is common to adopt a nonparametric viewpoint here and use a kernel density estimator $\widehat f_h(x)=n^{-1}\sum_{i=1}^nK_h(x-X_i)$, where the kernel $K$ is a unimodal, radially symmetric density, the bandwidth $h$ is a positive number and the notation $K_h(x)=K(x/h)/h^d$ represents the scaled kernel.

This paper explores a new methodology that arises as a blend of the two previous density-based clustering schemes. There are two possibilities for this blending. As a first option, noting that the kernel density estimator is a mixture density itself, one could try to apply mixture model clustering to such particular mixture density estimator. However, this makes little sense, since following the principles of mixture model clustering would lead us to declare that each of the $n$ ``mixture components" $K_h(x-X_i)$ forms a separate cluster, which besides contains a single data point ($X_i$). The second possibility, the other way round, involves applying the modal clustering methodology when the density estimate is obtained as the result of fitting a mixture model to the data, and indeed this makes perfect sense.

So, the main goal of this paper is to illustrate how the principles of modal clustering can be combined with mixture modeling. A related recent paper by \cite{S16} precisely shows how this can be done when high density clusters are understood as connected components of density level sets. Here, on the contrary, we focus on the notion of high density clusters as domains of attraction of the density modes, expanding on some ideas previously presented in \cite{Ch12}. Even if the two methodologies lead to very similar results in practice, the main advantage of the latter is that it is much simpler from a computational point of view, especially for high dimensional data.

The rest of the paper is organized as follows. In Section 2, both approaches, mixture model clustering and modal clustering, are compared to each other, and the pros and the cons of the two methodologies are exemplified through the analysis of synthetic and real data sets. In Section 3, two methods are introduced with the aim of producing a clustering from a modal point of view, but starting from a mixture model fit. Both methods rely on the use of the mean shift algorithm for normal mixture densities, and a new representation of this algorithm as a quasi-Newton optimization method is provided. Section 4 includes two more synthetic examples that show that it is possible to get close to the population modal clustering even if the starting point is a mixture density estimate. The paper finishes with a discussion section, posing some related open problems for future research.

\section{Mixture model clustering versus modal clustering}\label{sec:2}

\begin{figure}[t!]\centering

\begin{tabular}{@{}cc@{}}
\includegraphics[width=0.48\textwidth]{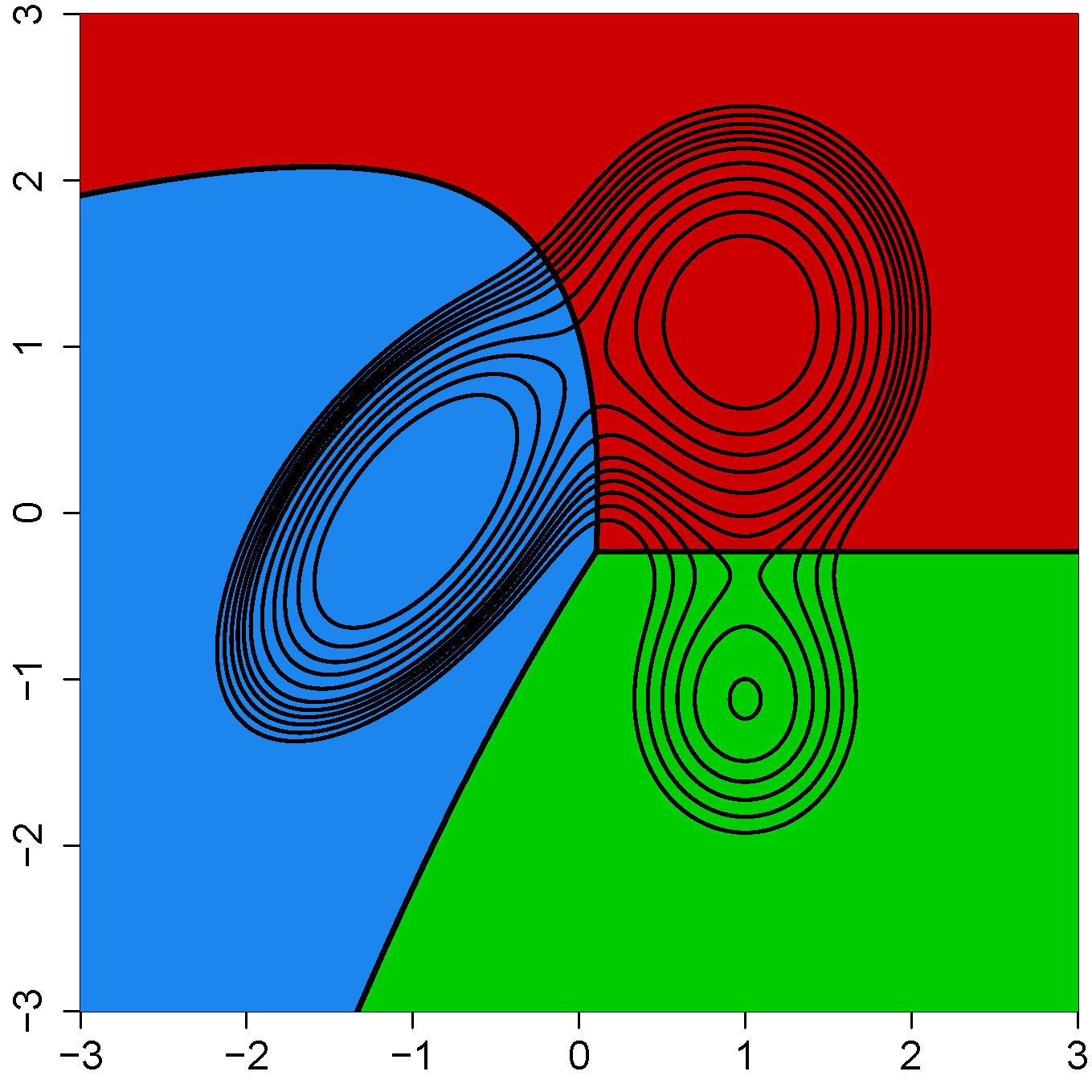}&\includegraphics[width=0.48\textwidth]{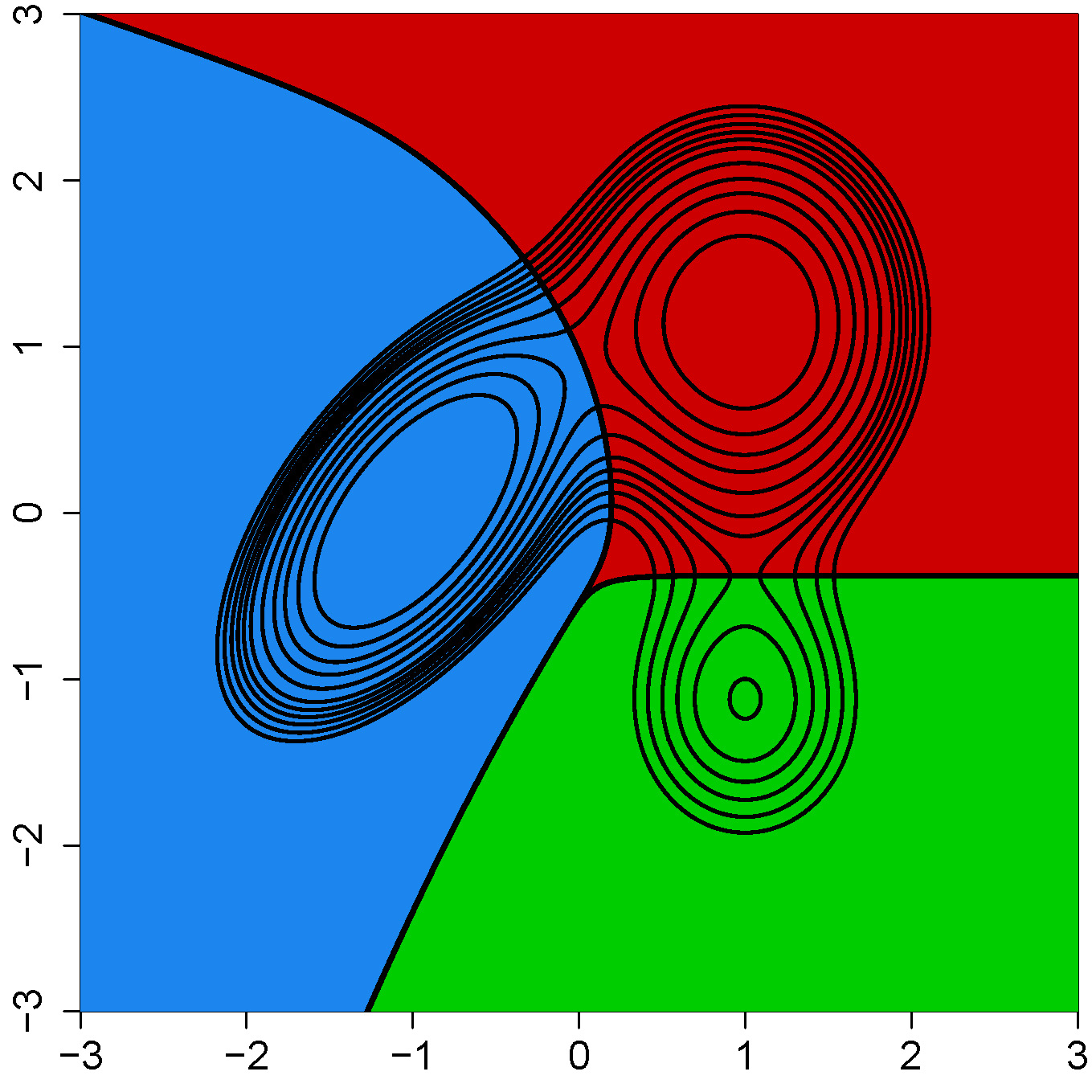}
\end{tabular}
\caption{Bivariate trimodal 3-component normal mixture density. Left, normal mixture model clustering. Right, modal clustering. The boundaries of the modal clusters are necessarily perpendicular to the contour lines, since they are flow lines of the gradient field.}
\label{fig:0}
\end{figure}

In principle, mixture model clustering and modal clustering aim at very different goals. It is not that one of these views is right and the other one is wrong. They just seek after different notions of cluster. And in fact, the two clusterings look very similar in ``non-problematic" situations, that is, when different mixture components correspond to different, well-separated unimodal distributions. As an example, Figure \ref{fig:0} shows a bivariate trimodal 3-component normal mixture density, with the population clusters depicted according to mixture model clustering (left) and modal clustering (right).

Differences arise precisely when there is not a one-to-one correspondence between mixture components and density modes. Consider the class $\mathcal P_0\equiv\mathcal P(f_0)$ of finite mixture distributions based on a fixed parametric distribution $f_0(\cdot|\theta)$.

\begin{figure}[t!]\centering

\begin{tabular}{@{}cc@{}}
\includegraphics[width=0.48\textwidth]{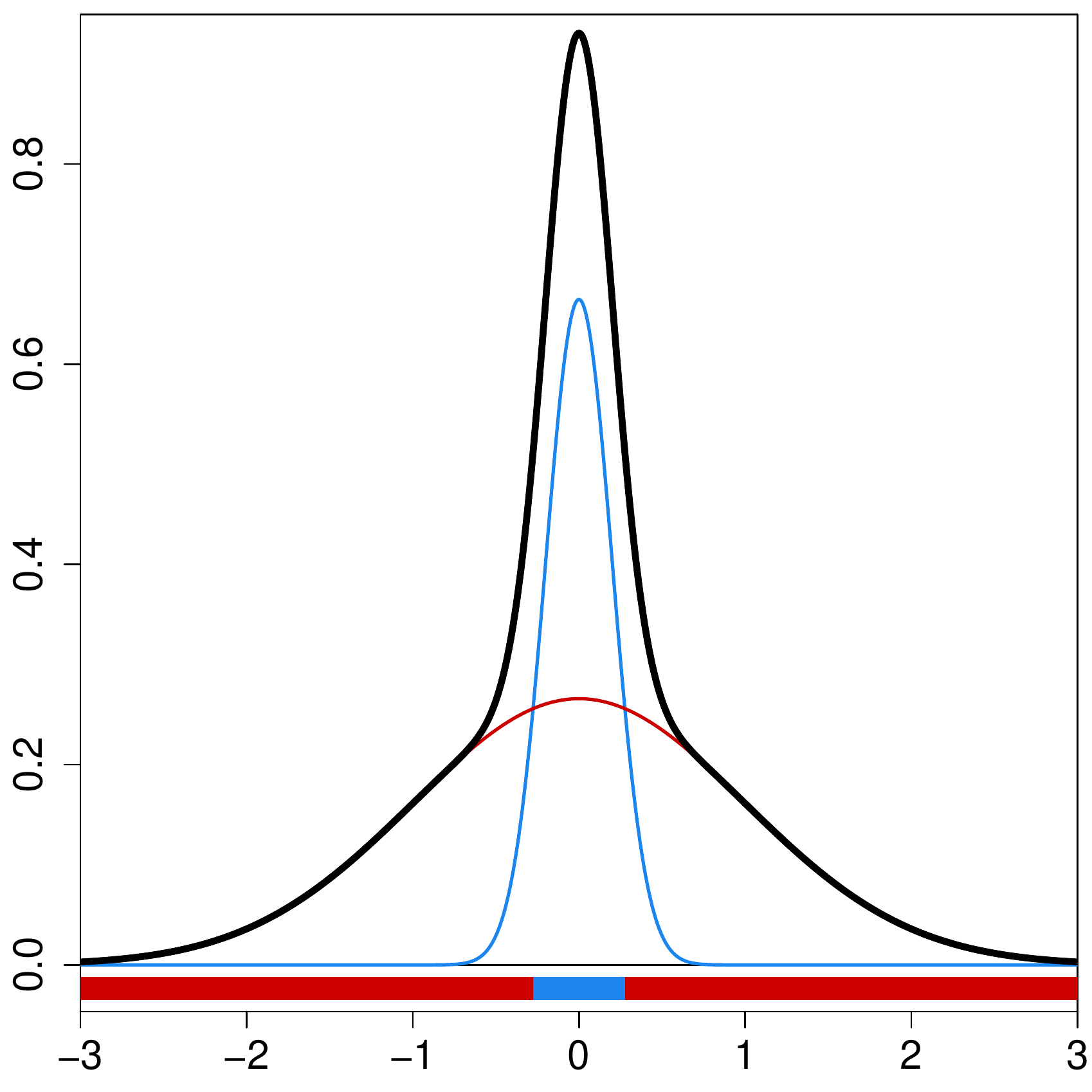}&\includegraphics[width=0.48\textwidth]{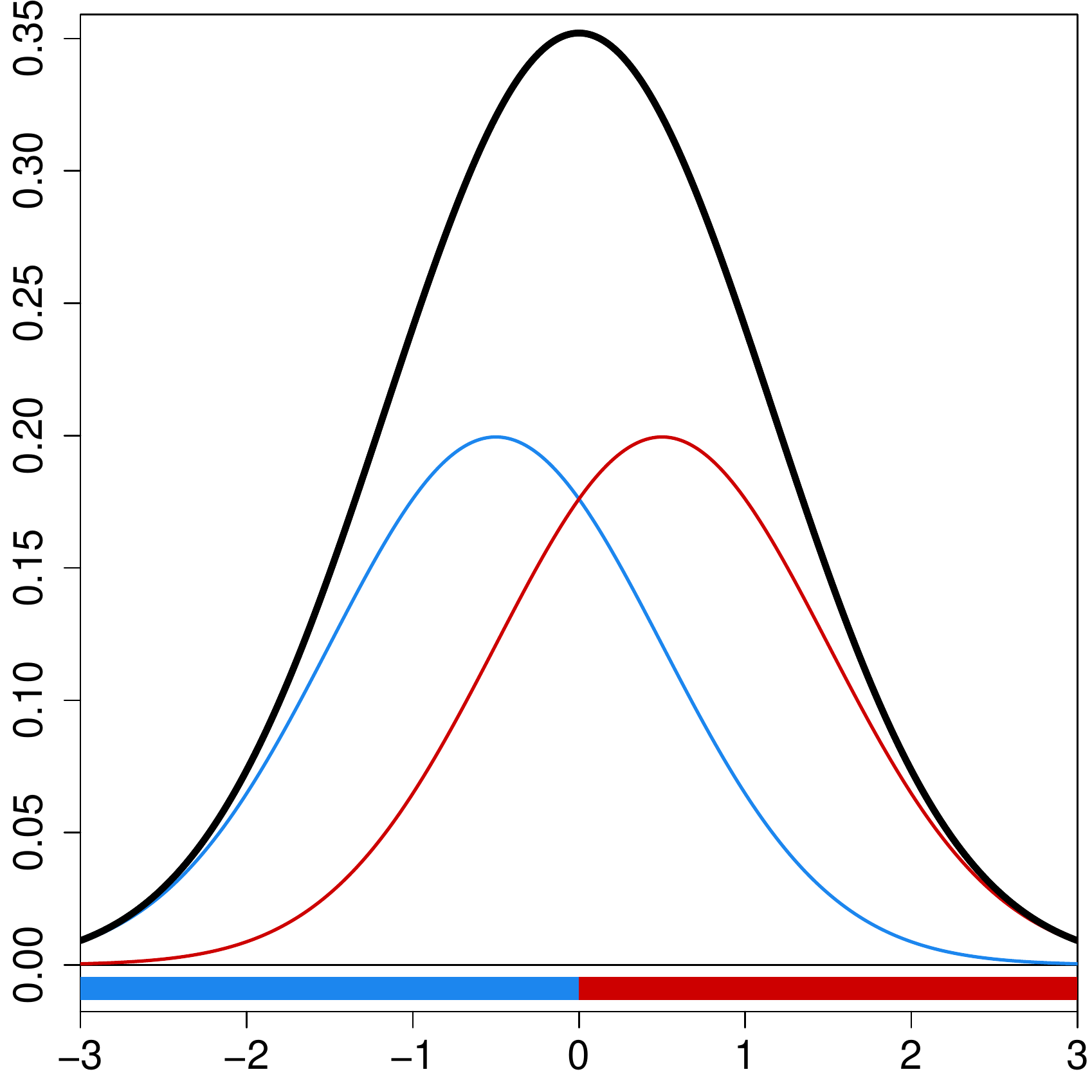}
\end{tabular}
\caption{Two examples of 2-component Gaussian mixture densities. The black line is the mixture density, the red and blue lines are the (weighted) density components, and the bottom line represents with the corresponding colors the mixture model clusters defined by the density components.}
\label{fig:1}
\end{figure}

If the true distribution of the data is indeed in $\mathcal P_0$, then mixture model clustering allows to distinguish more subtle situations than modal clustering. For instance, it notices when there are two populations with the same center but different dispersion, or even different centers, but close enough to result in a unimodal distribution. The left and right panels of Figure \ref{fig:1}, respectively, represent these two scenarios. Modal clustering, on the contrary, only observes the overall resulting density (not its components) and since the two components are not sufficiently separated, it cannot detect two separate groups and notices only one cluster in these distributions.

Nevertheless, since the class $\mathcal P_0$ is dense in the set of all density functions under the $L_1$ metric \citep[see, e.g.,][]{LB00}, in fact mixture modeling can be used in a nonparametric way to estimate any density, either belonging to $\mathcal P_0$ or not \citep{P94}. However, even if mixture modeling is thus useful for nonparametric density estimation, mixture model clustering should be used with caution when the true distribution of the data does not belong to the class $\mathcal P_0$ employed to estimate the density, because in this case, it may be misleading to identify mixture components with clusters \citep{H10}.

\begin{figure}[t!]\centering

\begin{tabular}{@{}cc@{}}
\includegraphics[width=0.4\textwidth]{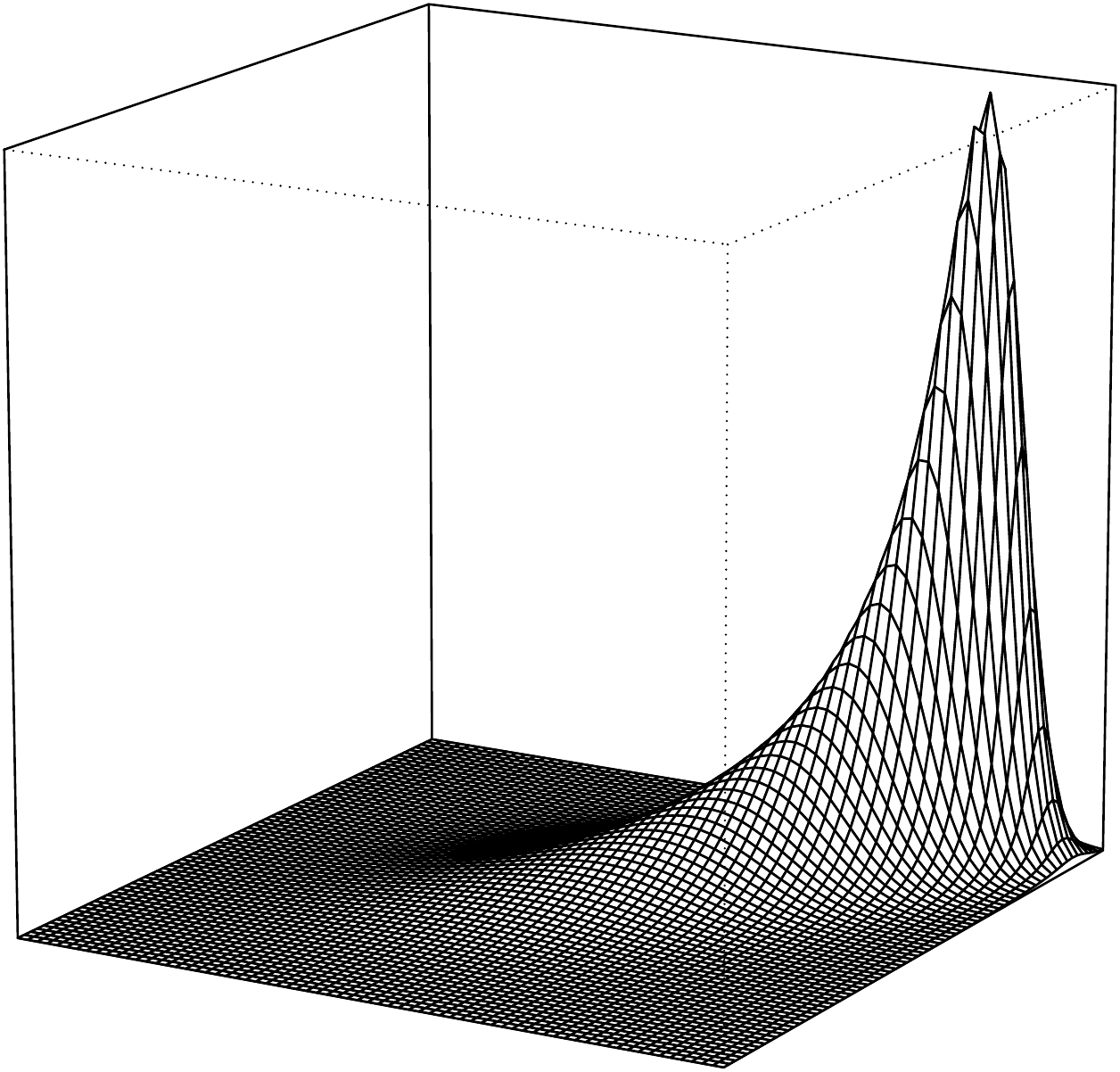}&\includegraphics[width=0.4\textwidth]{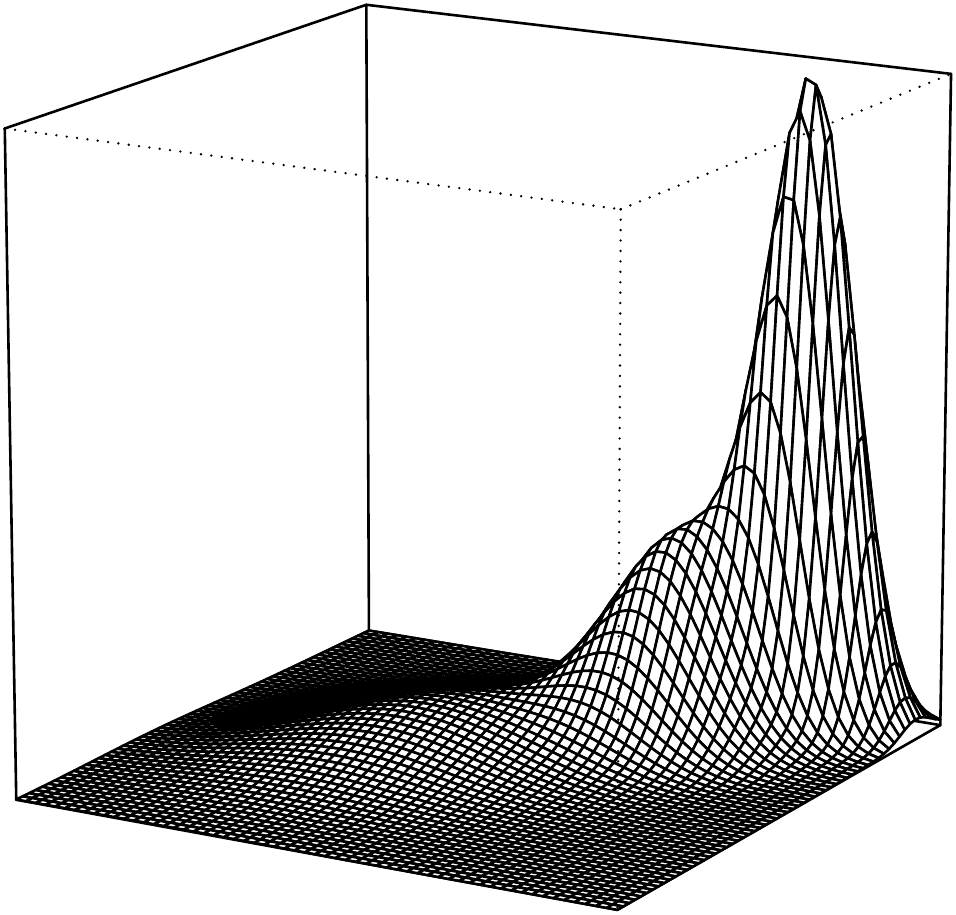}\\
\includegraphics[width=0.45\textwidth]{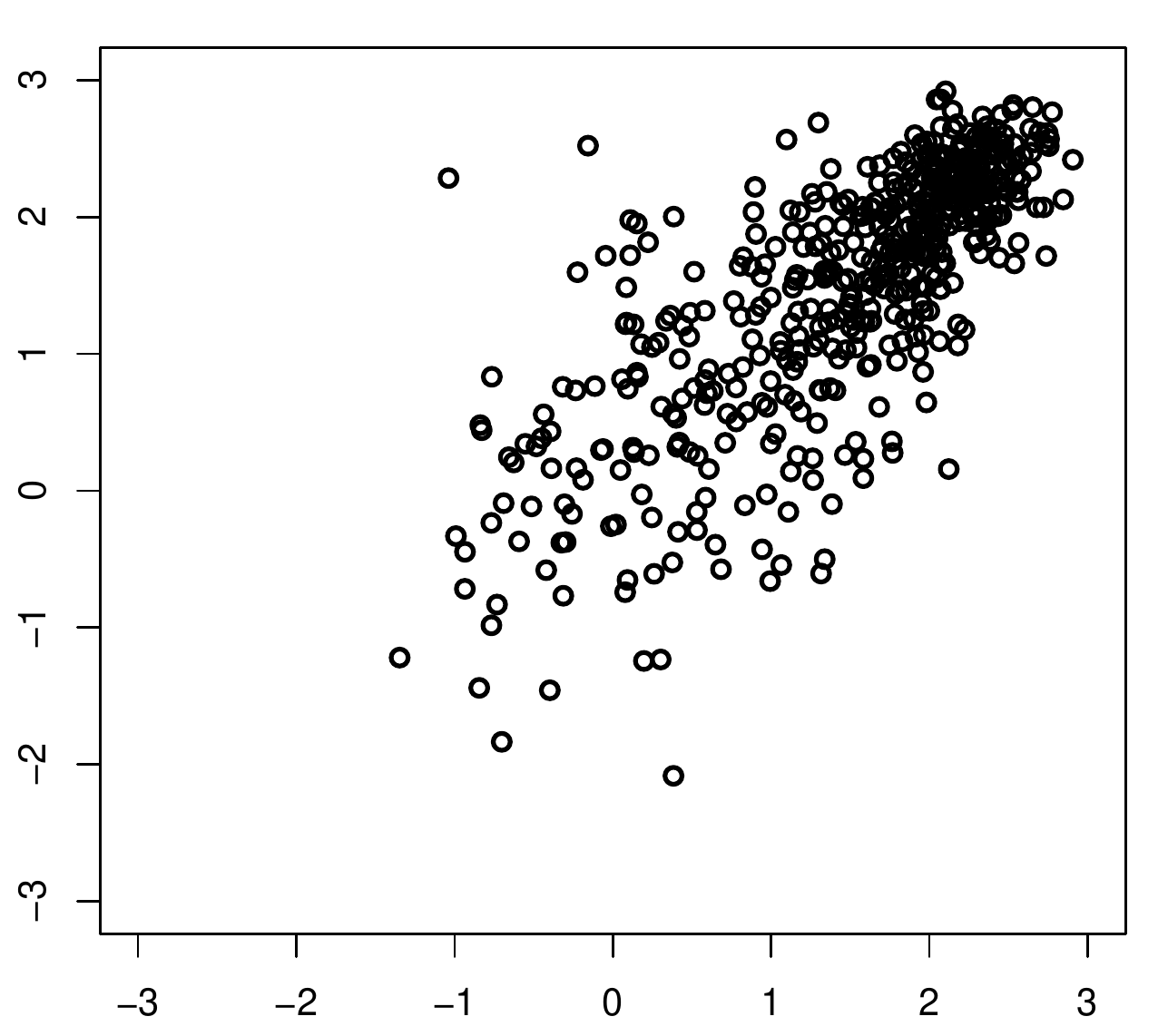}&\includegraphics[width=0.45\textwidth]{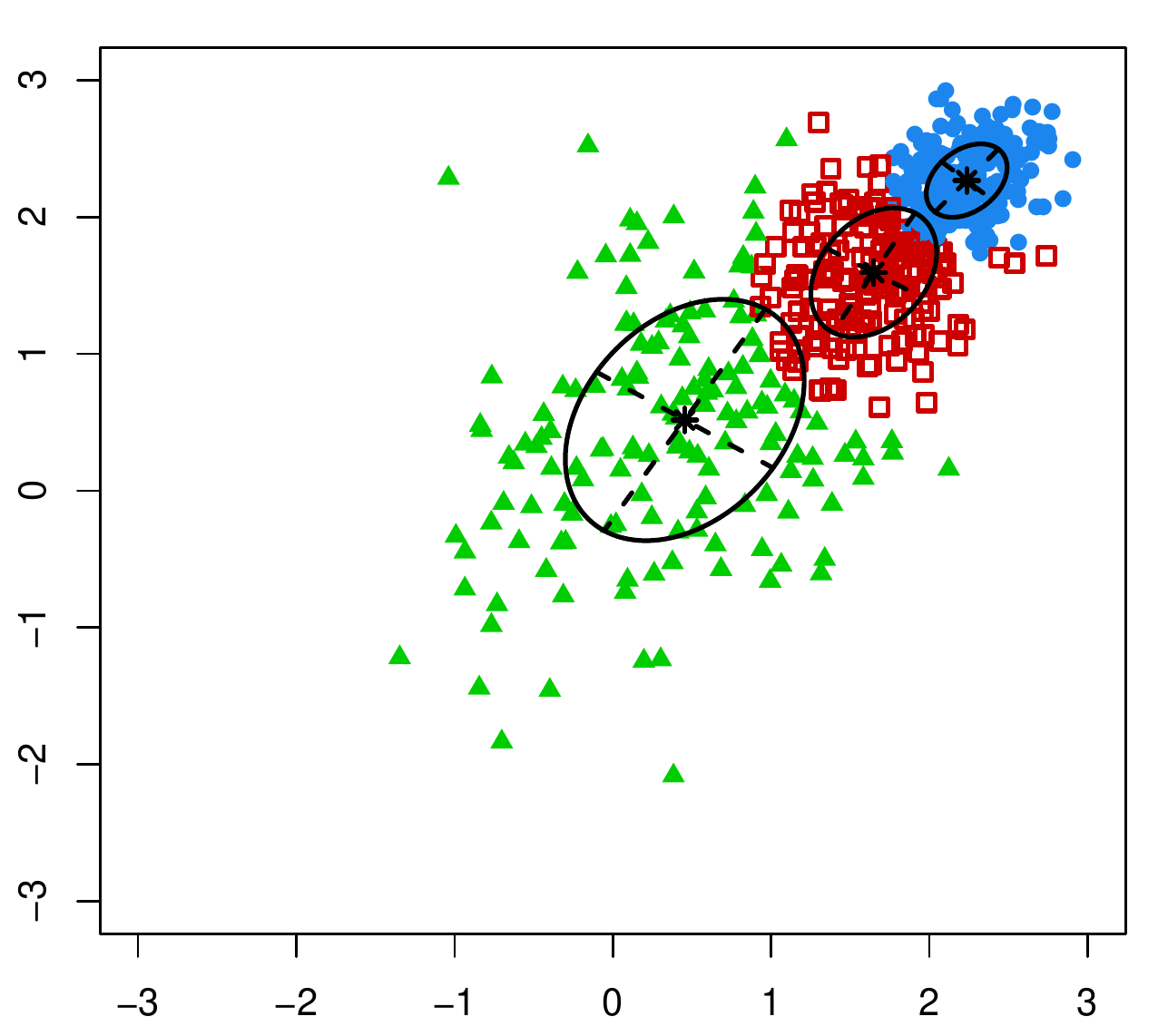}
\end{tabular}
\caption{Normal mixture model clustering applied to a skew-normal distribution. Top left, true distribution. Top right, normal mixture density estimate based on a sample of size $n=500$, represented in the bottom left picture. Bottom right: mixture model clustering, showing three clusters, as three normal components are needed to model the skew-normal distribution from this sample.}
\label{fig:2}
\end{figure}

For example, Figure \ref{fig:2} shows a bivariate skew-normal distribution and a normal mixture density estimate based on a sample of size $n=500$. The density estimate is reasonably close to the true density, but for that it is necessary to use a 3-component normal mixture. A blind application of mixture model clustering would result in a partition of the data into 3 clusters. But that seems quite artificial, since the same data are best fit using a single-component skew-normal mixture density \citep{L09}. In this case, adopting a modal clustering perspective after normal mixture density estimation yields one cluster, since the obtained density estimate is unimodal.

\begin{figure}[t!]\centering

\begin{tabular}{@{}cc@{}}
\includegraphics[width=0.4\textwidth]{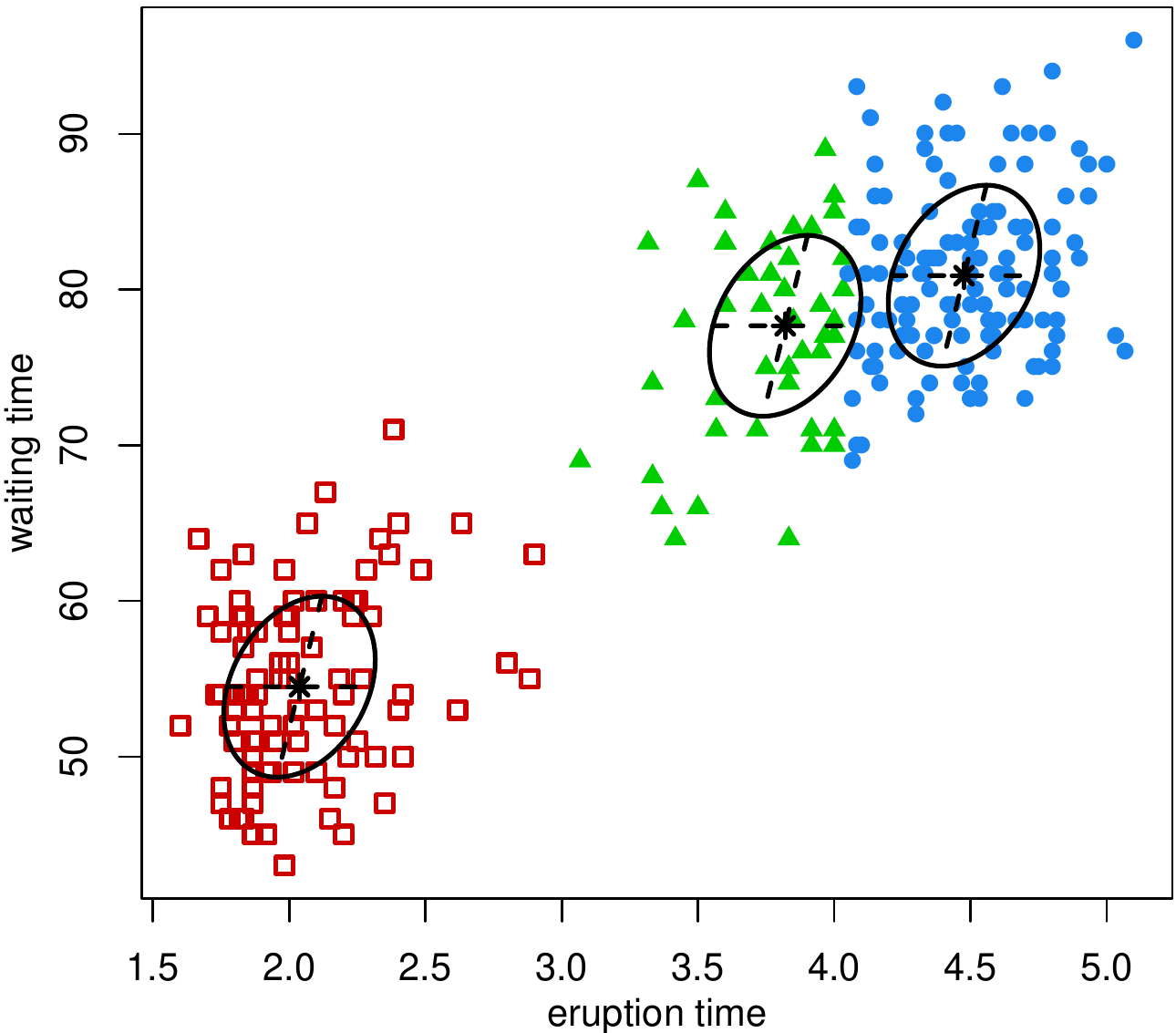}&\hspace{0.03\textwidth}\includegraphics[width=0.4\textwidth]{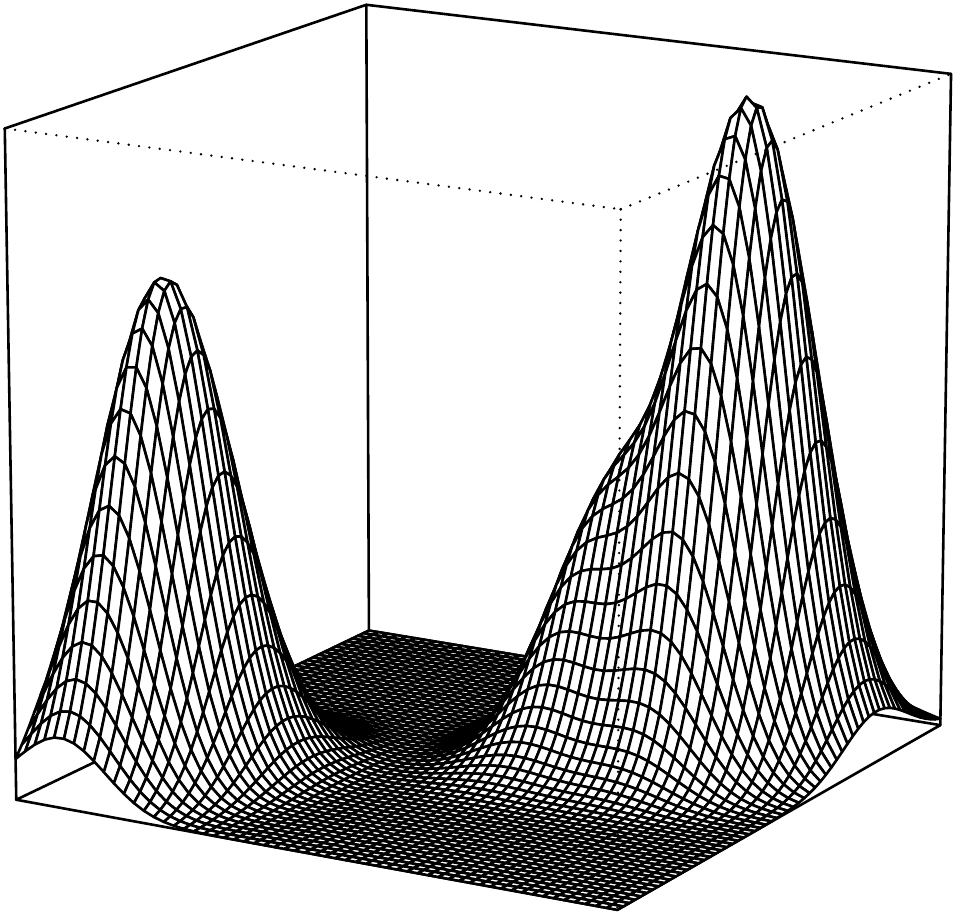}\\
\end{tabular}
\caption{Old Faithful data example. Left, data set and normal mixture clustering. Right, normal mixture density estimate.}
\label{fig:3}
\end{figure}

Another classical real data example where a similar situation occurs is the Old Faithful data set \citep{AB90}, which records the variables ``eruption time" and ``waiting time" regarding $n=272$ eruptions of the Old Faithful geyser in Yellowstone National Park. As noted in \cite{S16}, normal mixture density estimation results in a 3-component mixture for this data set, albeit the density estimate only shows two separate regions of high density (see Figure \ref{fig:3}). Indeed if the more general class of skew-normal mixture densities is employed, then the best fit is obtained using only two components.

A solution frequently proposed in the literature to amend this problem is to ``merge'' some components into the same cluster. The exhaustive paper by \citet{H10} provides an excellent review and comparison of the existing techniques to date, including some based on modality arguments. One of the proposals examined in this paper (Method 1 below) is indeed a further addition to this list, with a view towards simplicity and computational easiness.

\begin{figure}[t!]\centering

\begin{tabular}{@{}cc@{}}
\includegraphics[width=0.4\textwidth]{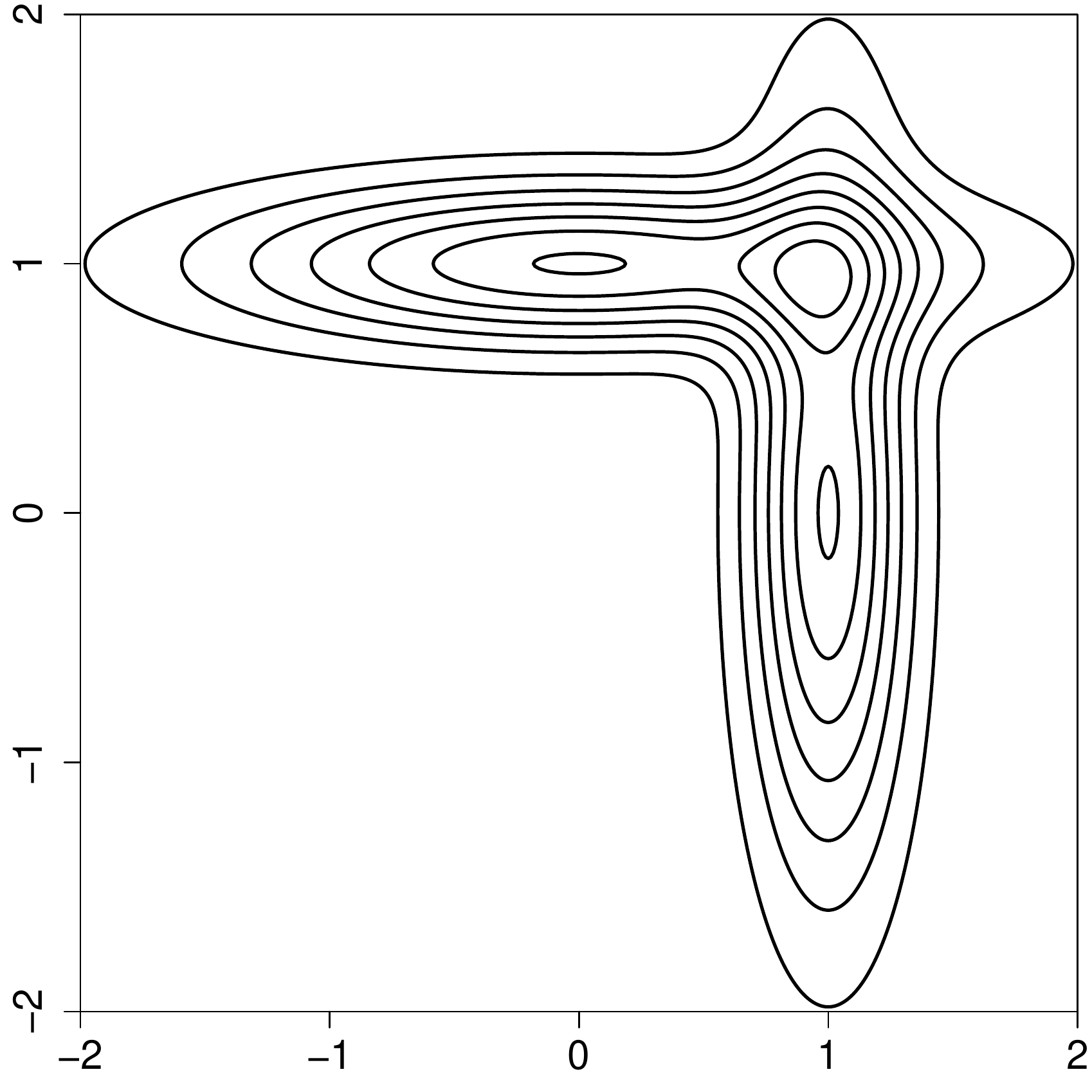}&\hspace{0.03\textwidth}\includegraphics[width=0.4\textwidth]{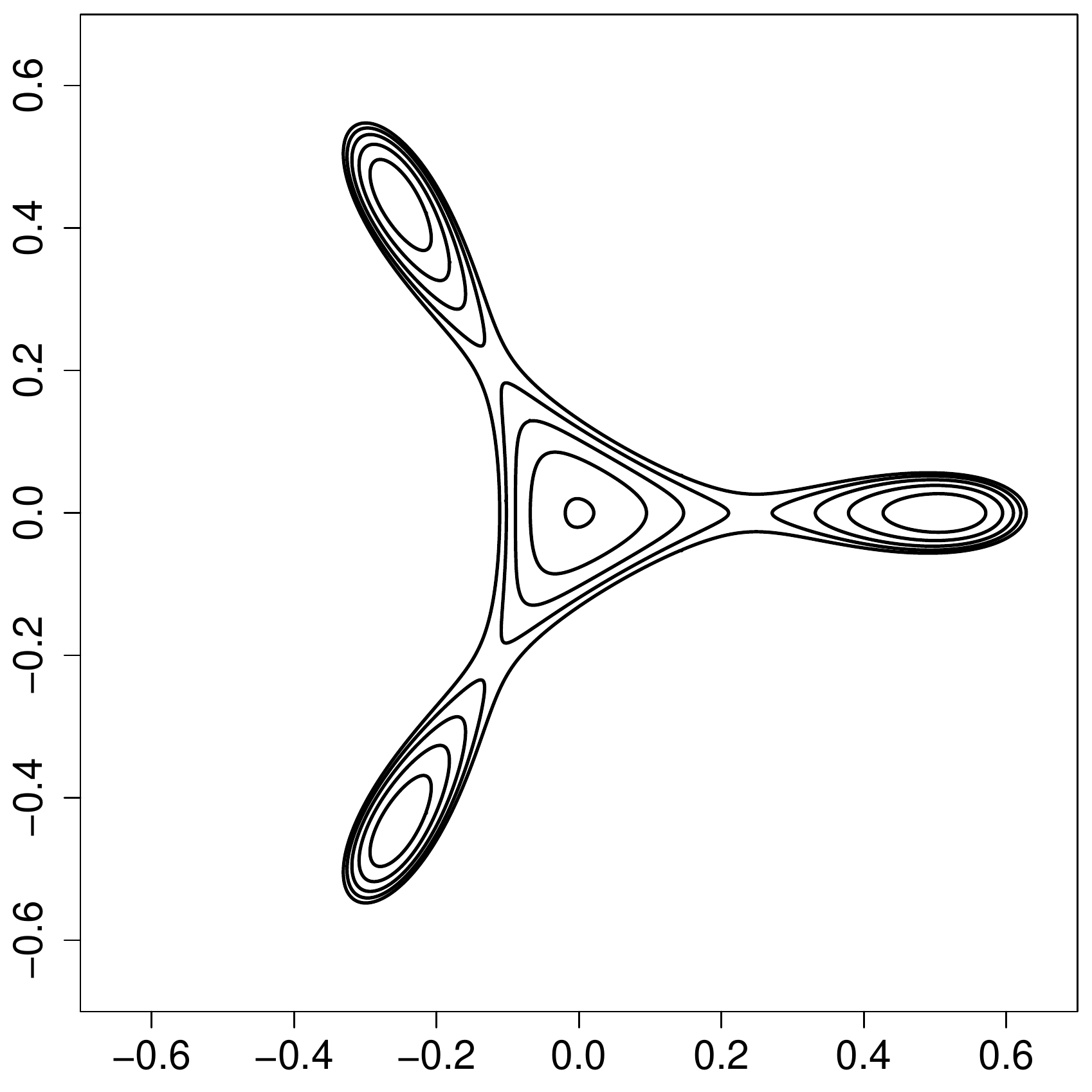}\\
\end{tabular}
\caption{Two normal mixture densities with more modes than components.}
\label{fig:4}
\end{figure}

In all the previous examples the disagreement between mixture model clustering and modal clustering is mainly due to the fact that the number of components is greater than the number of modes. This seems to be the most common case in practice. However, it is worth mentioning that the opposite situation may also occur; that is, a 2-component mixture may have more than two modes. These examples, however, appear to be quite less frequent in practice, and to produce them a thorough search of the mixture parameters is needed. Figure \ref{fig:4} shows a 2-component normal mixture density with three modes (left) and a 3-component isotropic normal mixture with four modes (right), as exposed in \cite{RL05} and \cite{CPW03b}, respectively. For the second example, this phenomenon occurs only for a small range of values of the scale parameter of the isotropic components \citep[see also][]{EFR13}.

For the class of normal mixture densities, the number of extra modes is quite controlled, since \cite{CPW03a} showed that for $d=1$ the number of modes cannot be larger than the number of components, and for $d>2$ \cite{RR12} showed that a $d$-variate 2-component normal mixture can have at most $d+1$ modes. However, for mixture densities based on a distribution different from the normal one, the situation can get much more complicated and more bizarre examples can be found, as shown in \cite{W03}. For this reason, we concentrate on normal mixture densities henceforth.

\section{Modal clustering after mixture modeling}

As announced in the previous sections, the goal of this paper is to illustrate how modal clustering can be applied after mixture model density estimation. For the reasons explained above, the density estimation step will be performed within the class of normal mixture distributions. That is, as a first step, by applying the expectation maximization (EM) algorithm to find the maximum likelihood estimates of the parameters and mixing weights, and the BIC to select the number of components \citep[see][for details]{FR02}, a density of the type $\widehat f(x)=\sum_{g=1}^{\widehat G}\widehat\pi_g\phi(x|\widehat\mu_g,\widehat \Sigma_g)$ is fitted to the data $X_1,\dots,X_n$. Here,
$$\phi(x|\mu,\Sigma)=(2\pi)^{-d/2}|\Sigma|^{-1/2}\exp\big\{-\tfrac12(x-\mu)^\top\Sigma^{-1}(x-\mu)\big\}$$
is the density of the $N(\mu,\Sigma)$ distribution.

In a second step, since the density estimate $\widehat f$ is a normal mixture density, we can make use of specifically designed mode-finding algorithms to investigate its modality features \citep[see][]{CP00}. This is the main difference with the proposal in \cite{S16}: while modal clustering based on connected components density level sets involve a high computational load, hindered by the need to compute the Delaunay triangulation of the data points, the mean-shift algorithm or one of its accelerated variants \citep{CP06,CP07} provides an efficient tool to perform modal clustering analysis from a normal mixture density estimate, even in high dimensions.

Depending on how we use the mean shift algorithm, it leads to two different modal clustering methods after mixture model density estimation: on one hand, a method for merging components, quite naive but extremely fast even for in high dimensional data; and, on the other hand, a clustering method that does not necessarily coincide with a merging of the mixture components.

\subsection{The non-isotropic mean shift algorithm as a quasi-Newton optimization method}\label{meanshift}

First we derive the formulation of the mean shift algorithm for a normal mixture density.

The gradient of a normal mixture density $f(x)=\sum_{g=1}^{G}\pi_g\phi(x|\mu_g,\Sigma_g)$ with respect to $x$ is easily shown to be
\begin{equation}
\mathsf D f(x)=\sum_{g=1}^G\pi_g\phi(x|\mu_g,\Sigma_g)\Sigma_g^{-1}(\mu_g-x),\label{Df}
\end{equation}
so to find the critical points of $f$ we would solve $\mathsf D f(x)=0$ for $x$ to obtain $x=T(x)$ with
\begin{equation}
T(x)=\Big\{\sum_{g=1}^g\pi_g\phi(x|\mu_g,\Sigma_g)\Sigma_g^{-1}\Big\}^{-1}\sum_{g=1}^G\pi_g\phi(x|\mu_g,\Sigma_g)\Sigma_g^{-1}\mu_g.\label{Teq}
\end{equation}
Thus, from any initial point $y_0\in\mathbb R^d$, a sequence $\{y_j\}_{j=1}^\infty$ to approximate the critical points numerically can be iteratively constructed by setting $y_{j+1}=T(y_j)$, as in \cite{CP00}. \citet{LHW07} showed, under mild conditions, that the sequence $\{y_j\}_{j=1}^\infty$ is convergent for any initial $y_0$ \citep[see also][and references therein]{AG15}.

The mean shift algorithm \citep{FH75} was initially motivated as a gradient clustering algorithm, by application of a (normalized) gradient ascent maximization algorithm to an arbitrary initial point $y_0$, to obtain a sequence $y_{j+1}=y_j+a_j \mathsf Df(y_j)/f(y_j)$ for some step size $a_j>0$. \cite{AMP16} showed not only that the resulting mean shift sequence converges to a mode of $f$, but also that the polygonal line defined by linear interpolation of two consecutive steps provides a consistent estimator of the flow lines of the density gradient field.

Next we show that the previous iterative scheme $y_{j+1}=T(y_j)$ can also be obtained as a quasi-Newton optimization method of the form $y_{j+1}=y_j+B_j\mathsf Df(y_j)/f(y_j)$ for a positive definite matrix $B_j$, which makes it closer to the original mean shift idea. Starting from (\ref{Df}) and considering the weights $w_g(x)=\pi_g\phi(x|\mu_g,\Sigma_g)/f(x)$, which are positive and add to one (in fact, $w_g(x)$ can be recognized as the a posteriori probability of the $g$-th mixture component, given $x$), reasoning as in \cite{Co03} it is clear that
$$\mathsf Df(x)/f(x)=\sum_{g=1}^Gw_g(x)\Sigma_g^{-1}\mu_g-\overline\Sigma(x)^{-1}x,$$
where $\overline\Sigma(x)=\{\sum_{g=1}^Gw_g(x)\Sigma_g^{-1}\}^{-1}$ is a weighted harmonic mean of the variance matrices of the normal mixture. Therefore, by taking $B_j=\overline\Sigma(y_j)$ it follows that $T(y_j)=y_j+B_j\mathsf Df(x)/f(x)$, and hence the iterative scheme $y_{j+1}=T(y_j)$ can also be seen as a quasi-Newton maximization algorithm.

In the isotropic case, multiplication by the matrix $B_j$ is simplified to multiplication by a constant $a_j$, so the procedure has the form of a gradient ascent algorithm, as in the original formulation of the mean shift algorithm.

\subsection{The two methods}

After fitting a normal mixture density $\widehat f(x)=\sum_{g=1}^{\widehat G}\widehat\pi\phi(x|\mu_g,\Sigma_g)$ to the data, there are two possibilities to obtain a modal clustering:

\subsubsection{Method 1: modal merging of mixture components}

The $\widehat G$ whole-space clusters obtained from the normal mixture fit are $\widehat{\mathcal C}_1,\dots, \widehat{\mathcal C}_{\widehat G}$, where
\begin{equation}
\widehat{\mathcal C}_g=\big\{x\in\mathbb R^d\colon\widehat\pi_g\phi(x|\widehat \mu_g,\widehat\Sigma_g)\geq\widehat\pi_j\phi(x|\widehat \mu_j,\widehat\Sigma_j),\ \forall j\neq g\big\}\label{mixclus}
\end{equation}
for $g=1,\dots,\widehat G$. If the goal is just to cluster the data set $X_1,\dots,X_n$, then each of this data points should be assigned to the cluster $\widehat {\mathcal C}_g$ where it belongs.

But, as it was illustrated in Section \ref{sec:2}, it could happen that two or more of these clusters represent in fact a single unimodal distribution. So in that case, from a modal clustering point of view, it would be advisable to merge into a single cluster all the components that give rise to the same unimodal distribution. This is easy to do with the naked eye, after plotting the resulting density estimate, if the data dimension is one or two. But for higher dimensional data it is necessary to have an automated algorithm to do so.

In \cite{H10}, two methods are proposed to achieve this goal, based on the concept of ridgeline introduced in \citet{RL05}. They consist in examining if every posible pair of fitted mixture components leads to a unimodal distribution or not (in the second method, a tuning parameter is introduced to allow merging two components even if the resulting mixture is not unimodal, provided the valley in the ridgeline that identifies the two modes is not too deep). A disadvantage of this methodology, as noted by the author, is that since the comparisons are made two by two components, it induces a hierarchical merging that may cause trouble when the algorithm is re-run again to look for further mergings.

From a slightly different point of view, notice that if the mixture of two or more components result in a unimodal density region, then their means belong to the domain of attraction of that same mode (because if one belonged to the domain of attraction of another mode, there should be a valley in the density separating it from the other components, and then their mixture could not be unimodal). Therefore, our proposal for merging based on modal clustering is to apply the mean shift method in Section \ref{meanshift} starting from each of the estimated component means $\widehat\mu_1,\dots,\widehat \mu_{\widehat G}$, and merge all the components whose estimated means converge to the same mode of $\widehat f$. In contrast with the methods based on the ridgeline, here the merging process is not pairwise nor hierarchical; rather, all the components are dealt with at the same time. As a result we obtain a new clustering $\widetilde{\mathcal C}_1,\dots,\widetilde{\mathcal C}_{\widehat M}$, where $\widehat M$ is the number of modes of $\widehat f$ and $\widetilde{\mathcal C}_m$ is made of the union of those clusters out of $\widehat{\mathcal C}_1,\dots, \widehat{\mathcal C}_{\widehat G}$ whose component means converge to the $m$-th mode of $\widehat f$. The whole process is summarized in Algorithm \ref{merging}.

\begin{algorithm}[h!t]
\caption{Modal merging of mixture components}\label{merging}
\SetKwInOut{Input}{Input}\SetKwInOut{Output}{Output}
\Input{data $X_1,\dots,X_n$}
\Output{clustering $\widetilde{\mathcal C}_1,\dots,\widetilde{\mathcal C}_{\widehat M}$, where $\widehat M$ is the number of estimated modes}
\begin{enumerate}
\item Fit a normal mixture $\widehat f$ to the data
\item Find the mixture component clusters $\widehat{\mathcal C}_1,\dots, \widehat{\mathcal C}_{\widehat G}$, defined as in (\ref{mixclus})
\item Run the mean shift algorithm with all the estimated component means $\widehat \mu_1,\dots,\widehat\mu_{\widehat G}$ as initial values
\item Build up $\widetilde{\mathcal C}_m$ as the union of those clusters out of $\widehat{\mathcal C}_1,\dots, \widehat{\mathcal C}_{\widehat G}$ whose component mean converges to the $m$-th mode of $\widehat f$, for $m=1,\dots,\widehat M$
\end{enumerate}
\end{algorithm}

The merging stage of the previous algorithm is simple and fast. Notice that once the mixture density is fitted, the mean shift algorithm only needs to be run with $\widehat G$ initial values, the estimated component means, which besides are typically not far from the modes \citep{RL05}. As a consequence, convergence is guaranteed after a reasonably small number of iterations, even in high dimensions.

\subsection{Method 2: modal clustering of the mixture density estimate}

The second proposal to perform modal clustering after normal mixture modeling is not inspired by component merging. Instead, in a more straightforward manner, it consists in obtaining the clusters as the domains of attraction of the modes of the fitted normal mixture $\widehat f$. To be precise, consider the integral curve $\widehat\gamma_x\colon\mathbb R\to\mathbb R^d$ to be the solution of the initial value problem
\begin{equation*}
\widehat\gamma_x'(t)=\mathsf D \widehat f\big(\widehat\gamma_x(t)\big),\  \widehat\gamma_x(0)=x.
\end{equation*}
Then, as in \cite{Ch15}, define the whole-space clustering associated to $\widehat f$ as the partition of the space with clusters $\widetilde{\mathcal D}_1,\dots, \widetilde{\mathcal D}_{\widehat M}$, given by
\begin{equation}\label{modal_clusters}
\widetilde{\mathcal D}_{m}=\big\{x\in\mathbb R^d\colon\lim_{t\to\infty}\widehat\gamma_x(t)=\widehat\tau_m\big\}
\end{equation}
for $m=1,\dots,\widehat M$, where $\widehat \tau_1,\dots,\widehat \tau_{\widehat M}$ are the modes of $\widehat f$. In practice, again we use the mean shift algorithm described in Section \ref{meanshift}, and its consistency as an estimator of the gradient flow lines \citep{AMP16}, to approximate $\lim_{t\to\infty}\widehat\gamma_x(t)\approx\lim_{j\to\infty} y_j$, where $y_{j+1}=\widehat T(y_j)$ with initial $y_0=x$, and $\widehat T$ the same as in (\ref{Teq}), but with the parameters replaced by their estimates. See Algorithm \ref{mcmde}.

\begin{algorithm}[h!t]
\caption{Modal clustering with a mixture density estimate}\label{mcmde}
\SetKwInOut{Input}{Input}\SetKwInOut{Output}{Output}
\Input{data $X_1,\dots,X_n$}
\Output{clustering $\widetilde{\mathcal D}_1,\dots,\widetilde{\mathcal D}_{\widehat M}$, where $\widehat M$ is the number of estimated modes}
\begin{enumerate}
\item Fit a normal mixture $\widehat f$ to the data
\item Use the iterative process $y_{j+1}=\widehat T(y_{j})$ to approximate the modal clusters $\widetilde{\mathcal D}_1,\dots, \widetilde{\mathcal D}_{\widehat M}$ defined as in (\ref{modal_clusters})
\end{enumerate}
\end{algorithm}

This second option is more meaningful from a modal clustering perspective, since instead of merging mixture component clusters, it is based on directly identifying the modal clusters of $\widehat f$. Yet, another difference is that it could have a higher computational cost. For instance, if the goal (as usual) is to cluster only the data $X_1,\dots,X_n$, then the mean shift algorithm has to be run with all this data points in the role of the initial value $y_0$, whereas for Algorithm \ref{merging} the mean shift procedure is applied only to the $\widehat G$ estimated component means as initial values, and in practice we usually have $\widehat G\ll n$.

\subsection{Comparing the two methods}

As shown in Figure \ref{fig:0} in population terms, mixture modal clustering and modal clustering result in different whole-space partitions, even if the number of components equals the number of modes. These partitions, however, are usually similar, so we can expect to observe an analogous phenomenon when working with a real data set.

Here, for illustration, we compare the results of Method 1 and Method 2 on a real data set. We use part of a big data set introduced in the study \cite{Bal07} of the Graft-versus-Host disease (GvHD). Different techniques to analyze these data have been proposed \citep{DCKW08,LBG08,LHL14}, including a method to merge normal mixture components \citep{Bal10}.

This data set contains measures obtained from flow cytometry techniques applied on two groups of patients, one group suffering the disease and a control group. We focus on four biomarker variables, namely, CD4, CD8$\beta$, CD3, and CD8, included in the package {\tt mclust} in R \citep{FRS16}. For the sake of brevity we use only the data from the GvHD-positive patients. Moreover, following \cite{LBG08} a sub-sample called CD$3^+$ is extracted by taking only the data assigned to components whose mean in the CD3 dimension is above 280 in a preliminary mixture fit. Finally, the variables CD4 and CD8$\beta$ corresponding to the resulting sub-sample are used to build a clustering using Methods 1 and 2 as described above.

\begin{figure}[t!]\centering

\begin{tabular}{@{}cc@{}}
\includegraphics[width=0.45\textwidth]{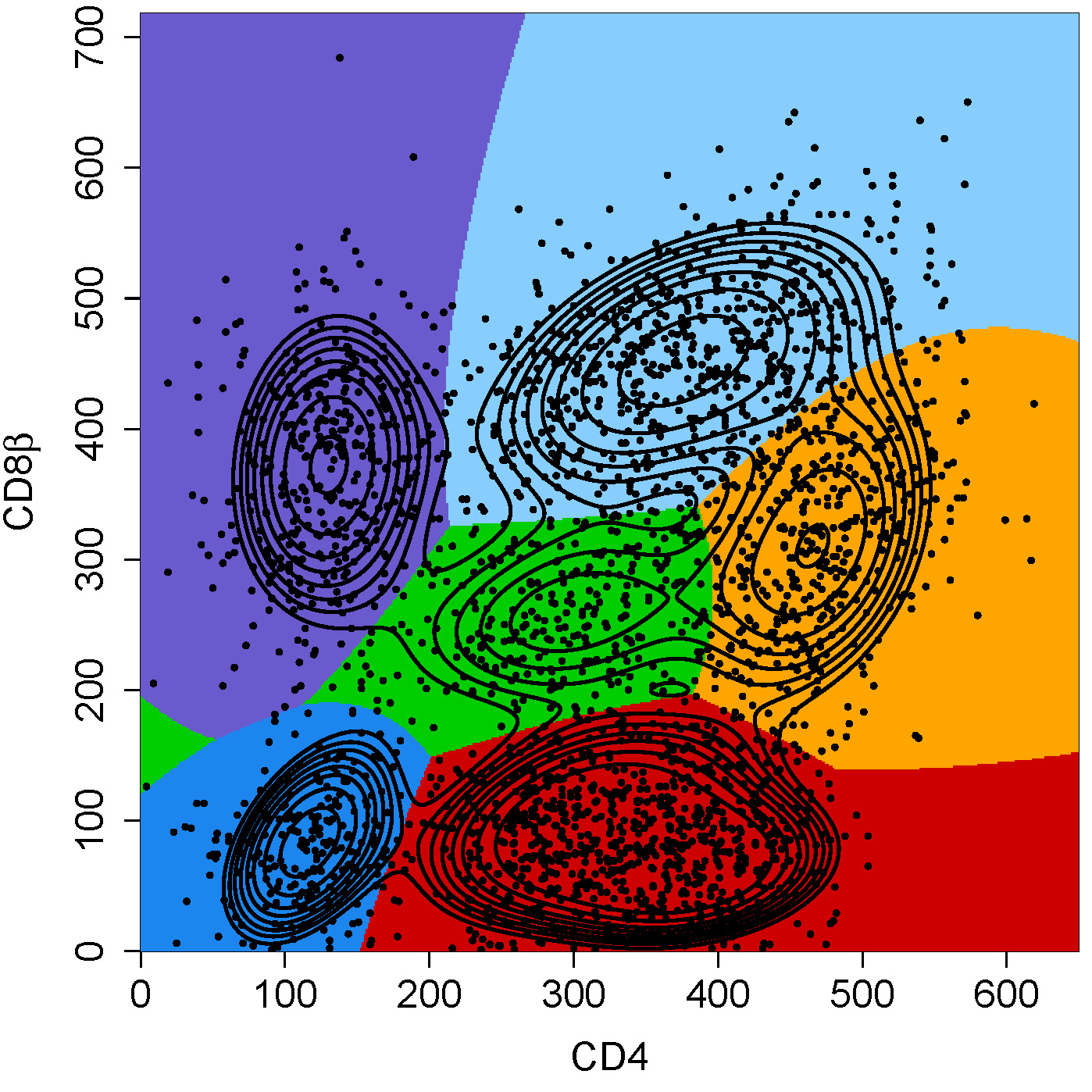}&\hspace{0.03\textwidth}\includegraphics[width=0.45\textwidth]{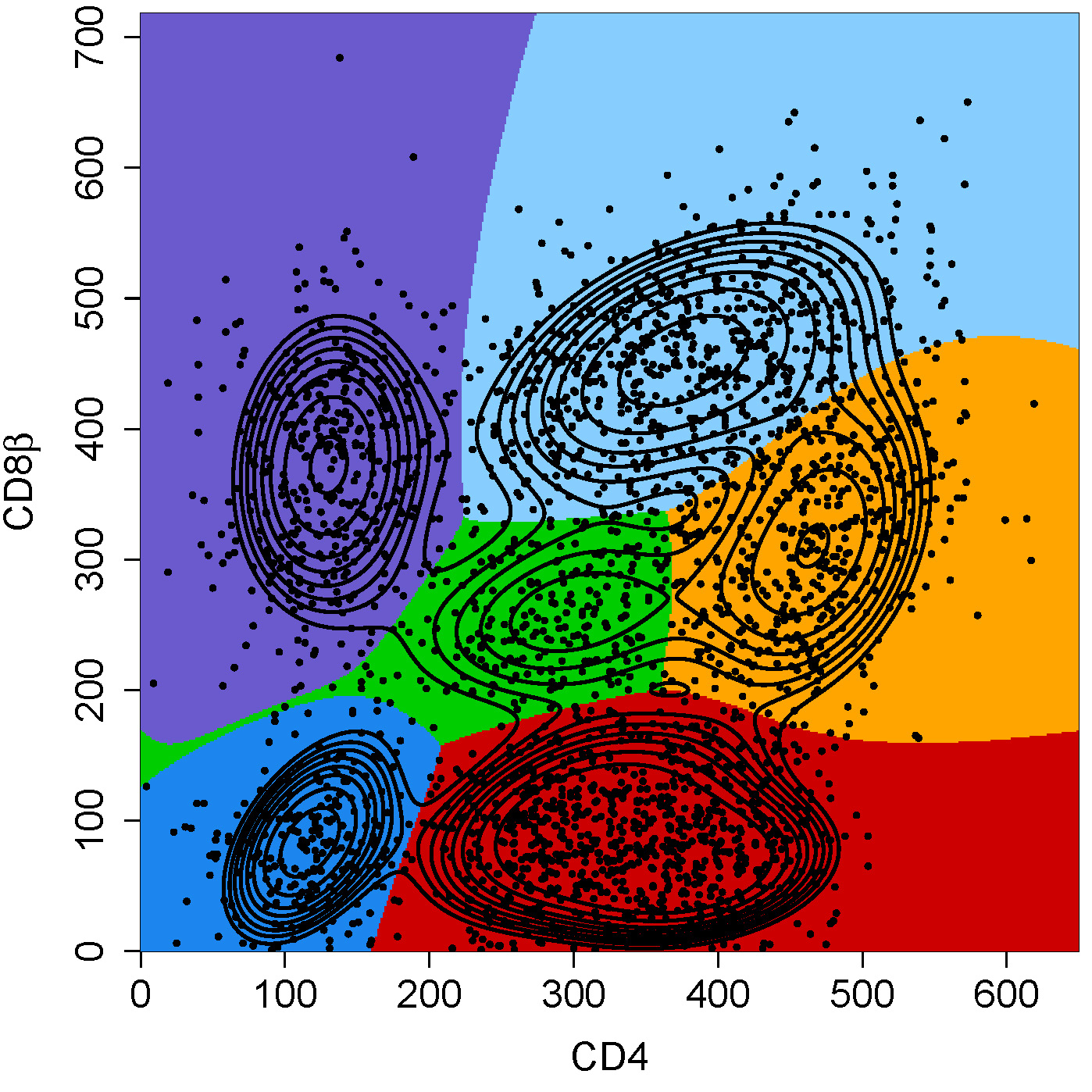}
\end{tabular}
\caption{GvHD data set. Left, normal mixture fit with modal merging. Right, modal clustering on the normal mixture fit.}
\label{fig:6}
\end{figure}

Figure \ref{fig:6} shows the result of the application of the two methods. The normal mixture fit obtained after the EM-BIC algorithm has 7 components and 6 modes. These 6 modes are in concordance with the manual analysis of \citet{Bal07}, who suggested that the CD$3^+$ cells could be divided into 6 cell sub-populations. Two of those 7 components are needed to approximate the bottom rightmost modal group, and Method 1 effectively merges these two components to obtain the clustering shown on the left picture of Figure \ref{fig:6}. If, instead, Method 2 is applied to obtain the modal clustering associated to the normal mixture fit, then the result is the clustering shown on the right picture of Figure \ref{fig:6}. Again, as anticipated above, both clusterings look very similar, with the most noteworthy visual difference being that the boundaries of the modal clustering are perpendicular to the density contour lines; this is something that does not occur for the modal merging solution.

\section{Additional examples}

Finally, we provide further illustration of the practical performance of the proposed methods by studying two additional synthetic examples where the population modal clustering (which is fully available in this simulation setting) differs by much from the mixture model clustering solution. In spite of that disagreement, performing modal clustering after a normal mixture fit is still able to recover the modal structure of the distribution in these cases.

\subsection{Overlapping components}

For the first simulated example we drew a sample of size $n=2000$ from the normal mixture with overlapping components introduced in \cite[][Section 4.1]{Bal10}. This is a 6-component normal mixture with 4 modes. Two of the modes are associated with two corresponding unimodal distributions with elipsoidal contours, while the other two modes are related to two, less usual, unimodal distributions with cross-shaped contours. The mixing proportions, means and variances defining this distribution, as extracted from \citet[][Appendix A.2]{B10}, are
\begin{gather*}
\pi_1=\pi_2=\pi_3=\pi_4=0.2,\,\pi_5=\pi_6=0.1,\\
\mu_1=(0,0),\,\mu_2=(8,5),\,\mu_3=\mu_4=(1,5),\,\mu_5=\mu_6=(8,0),\\
\Sigma_1=RAR^\top,\,\Sigma_2=R^\top AR,\,\Sigma_3=B,\,\Sigma_4=A,\,\Sigma_5=B,\,\Sigma_6=A,
\end{gather*}
where $A=\bigl(\begin{smallmatrix}1&0\\0&0.1\end{smallmatrix}\bigr)$, $B=\big(\begin{smallmatrix}0.1&0\\0&1\end{smallmatrix}\big)$ and $R=\frac12\big(\begin{smallmatrix}1&-\sqrt{3}\\\sqrt{3}&1\end{smallmatrix}\big)$.

As in \cite{Bal10}, only the results for one particular sample from this distribution are shown here for illustration purposes. However, these results represent in a sense the average situation arising from this distribution. For instance, the number of components fitted by the BIC for this particular sample was $\widehat G=9$, and the average number of components along 100 simulations runs of the same setup was 8.73; see Table \ref{tab:1}.

\begin{table}[t!]
\centering
\begin{tabular}{l||cccccccc}
$\widehat G$& 5&  6& 7 & 8& 9&10&11&12 \\\hline
Freq.&0.02&0.06&0.09&0.26&0.25&0.23&0.07&0.02
\end{tabular}
\caption{Distribution of $\widehat G$ selected by the BIC along 100 simulation runs from the normal mixture distribution with overlapping components ($n=2000$).}
\label{tab:1}
\end{table}

\begin{figure}[t!]\centering

\begin{tabular}{@{}cc@{}}
\includegraphics[width=0.4\textwidth]{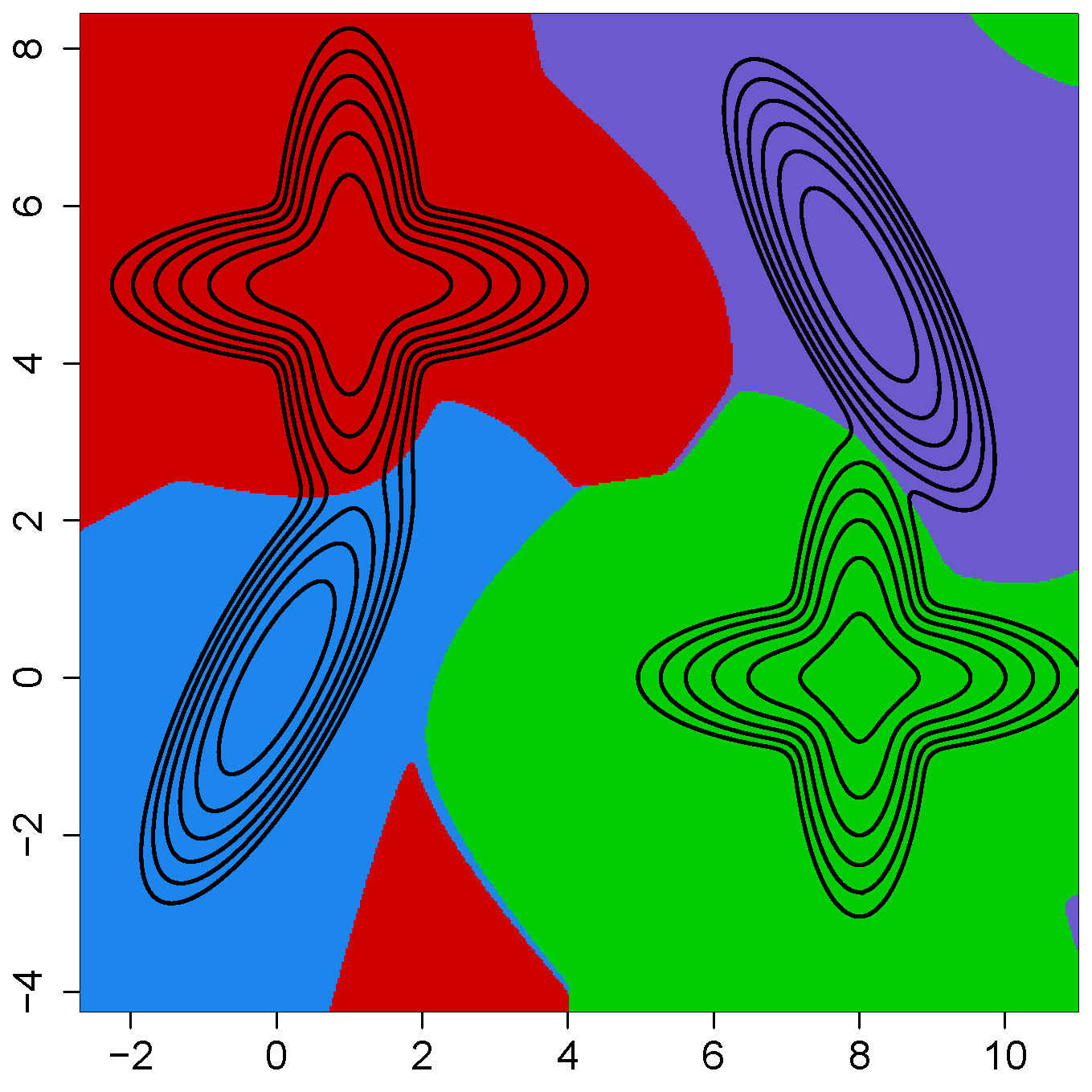}&\hspace{0.03\textwidth}\includegraphics[width=0.4\textwidth]{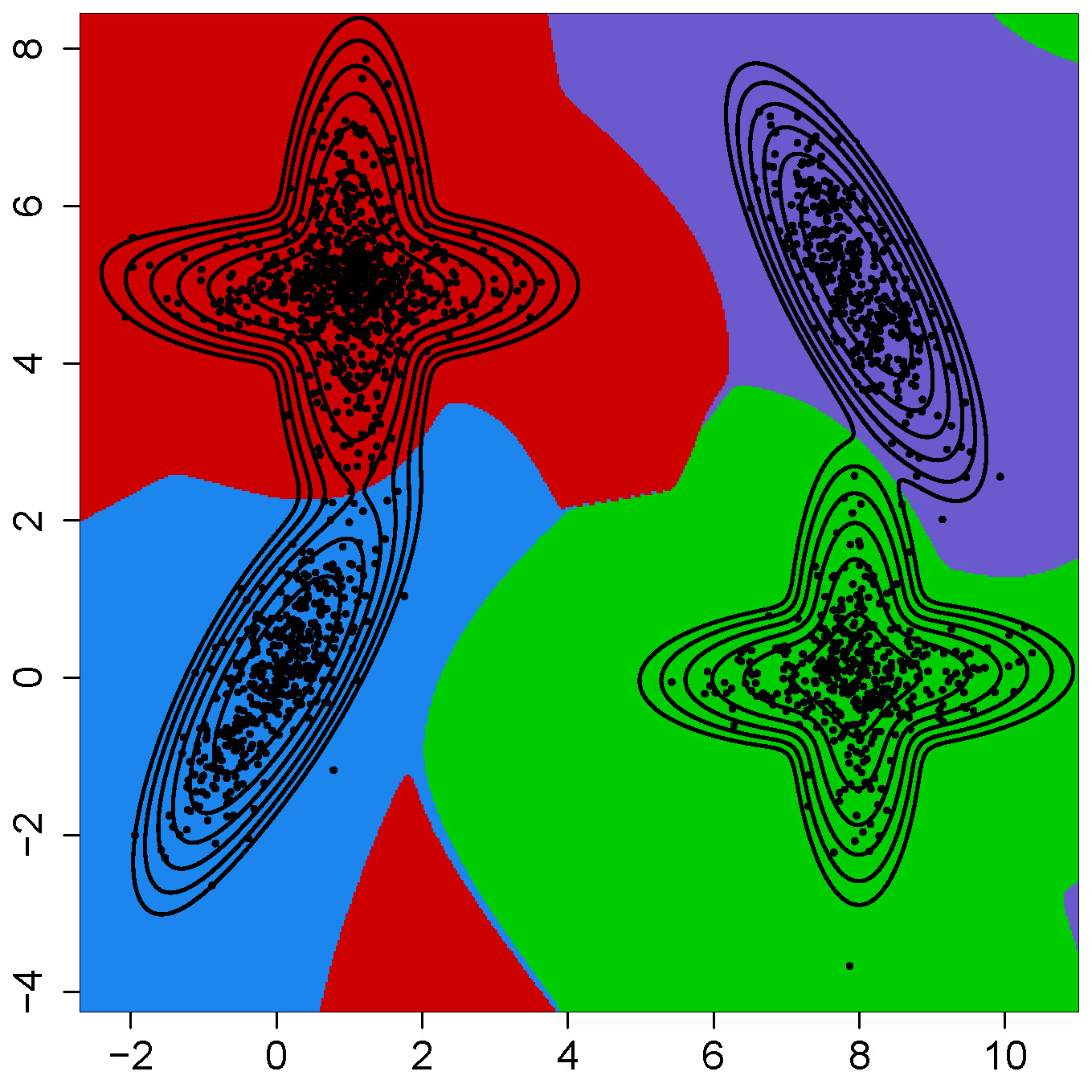}\\
\includegraphics[width=0.4\textwidth]{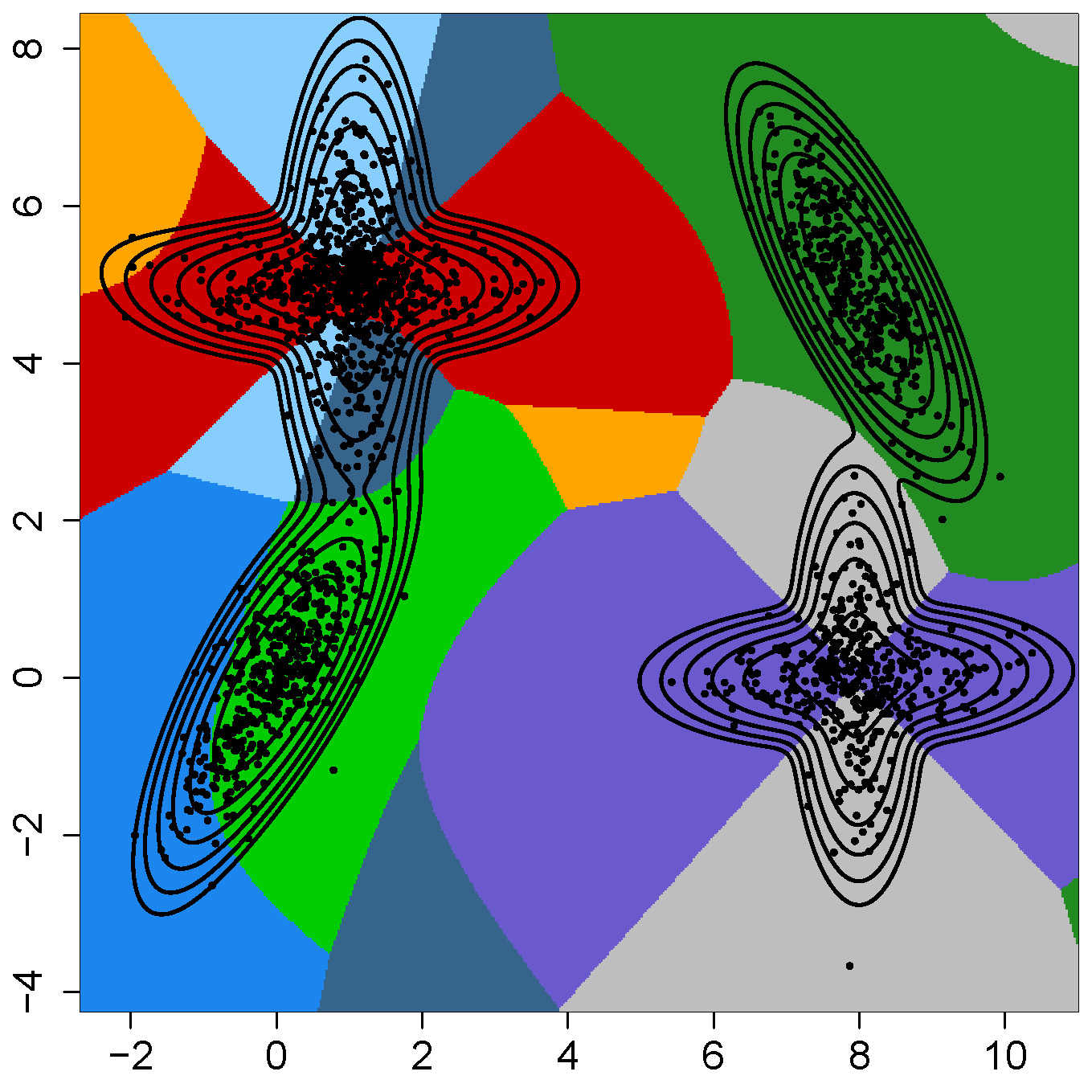}&\hspace{0.03\textwidth}\includegraphics[width=0.4\textwidth]{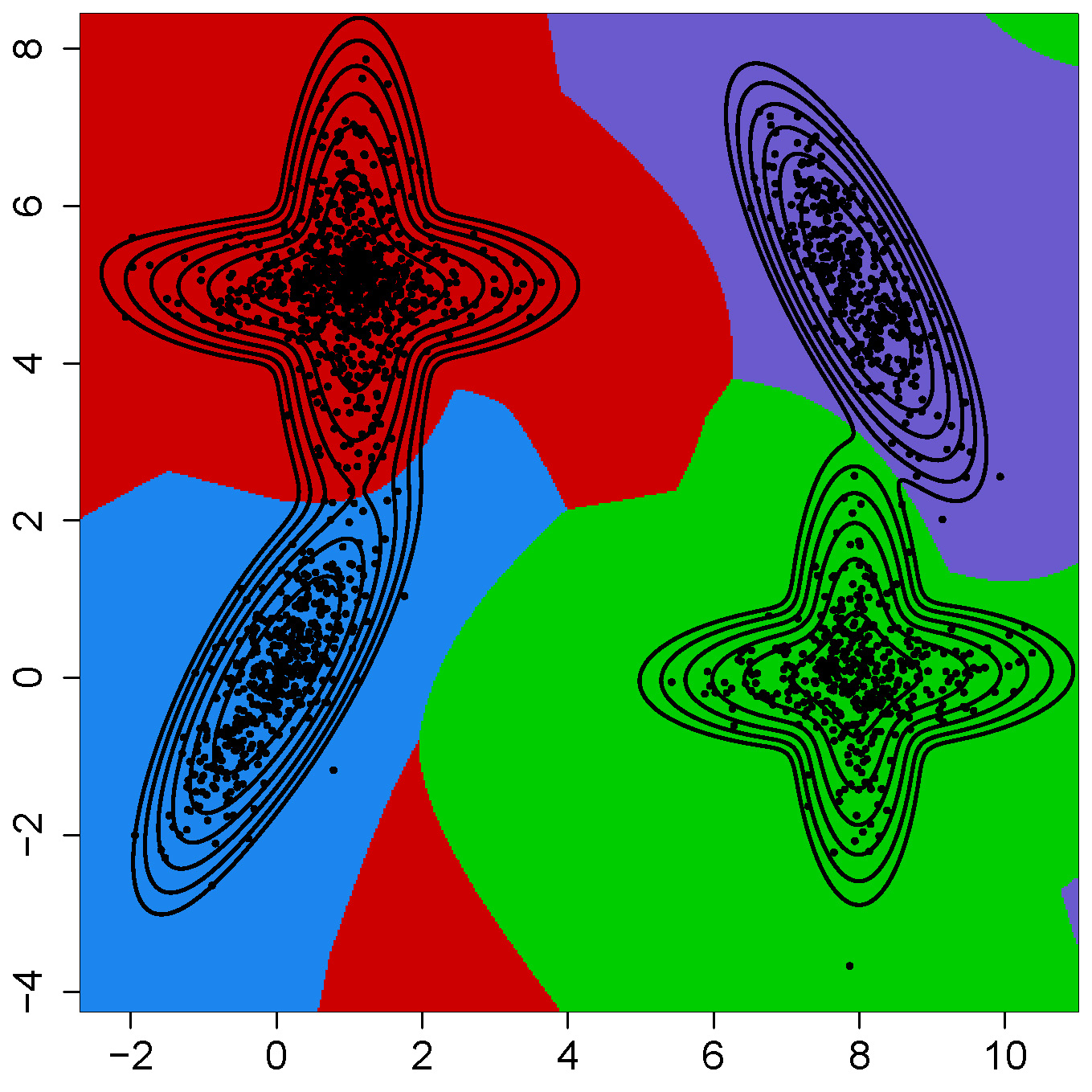}
\end{tabular}
\caption{Normal mixture with overlapping components. Top left, true modal clustering and true density contours. Top right, modal clustering on the normal mixture fit using Method 2. Bottom left, mixture model clustering with $G=9$ components. Bottom right, modal merging of the mixture components using Method 1.}
\label{fig:7}
\end{figure}

As the true distribution is known in this example, it is possible to obtain the true modal clustering, as defined in (\ref{modal_clusters}), but with respect to the true parameters. This is shown in the top left plot of Figure \ref{fig:7}. Top right picture shows the data points, the normal mixture density estimate and the modal clustering obtained through Method 2. It is visually apparent that the mixture fit produces a quite accurate density estimate. The bottom left picture of Figure \ref{fig:7} shows the clustering into 9 groups resulting from identifying clusters with mixture components, and the bottom right picture depicts how merging these 9 groups leads to 4 clusters after applying Method 1. Again, the two methods produce different but very similar clusterings, which in turn are also very close to the true population goal.

\subsection{Broken ring distribution}

As a second example we explore a distribution with five modal clusters, the broken ring distribution described in \citet[][Section 5.1]{CD13}. This distribution has a spherical bump in the middle, surrounded by four crescent-shaped clusters with different orientation.

\begin{figure}[t!]\centering

\begin{tabular}{@{}cc@{}}
\includegraphics[width=0.4\textwidth]{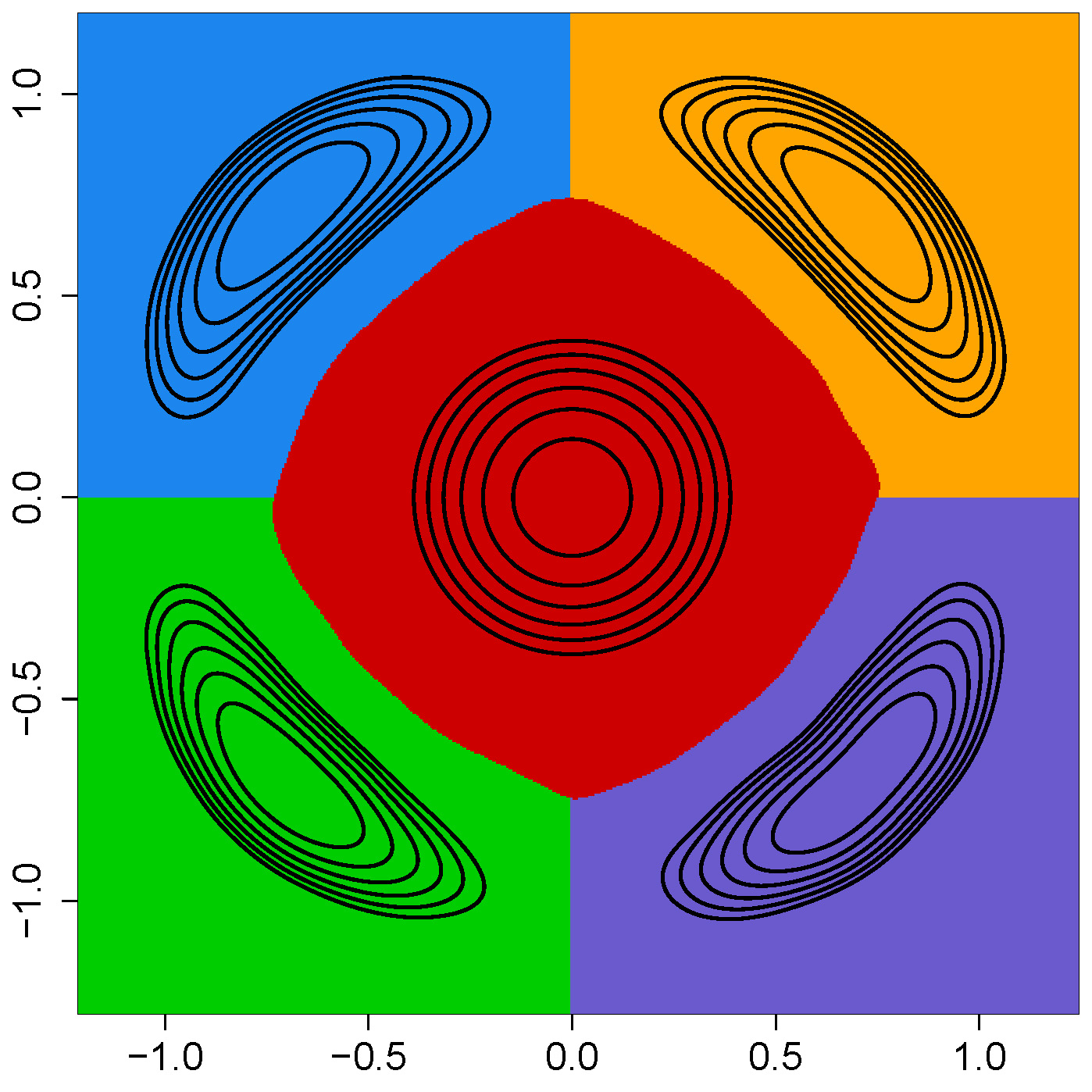}&\hspace{0.03\textwidth}\includegraphics[width=0.4\textwidth]{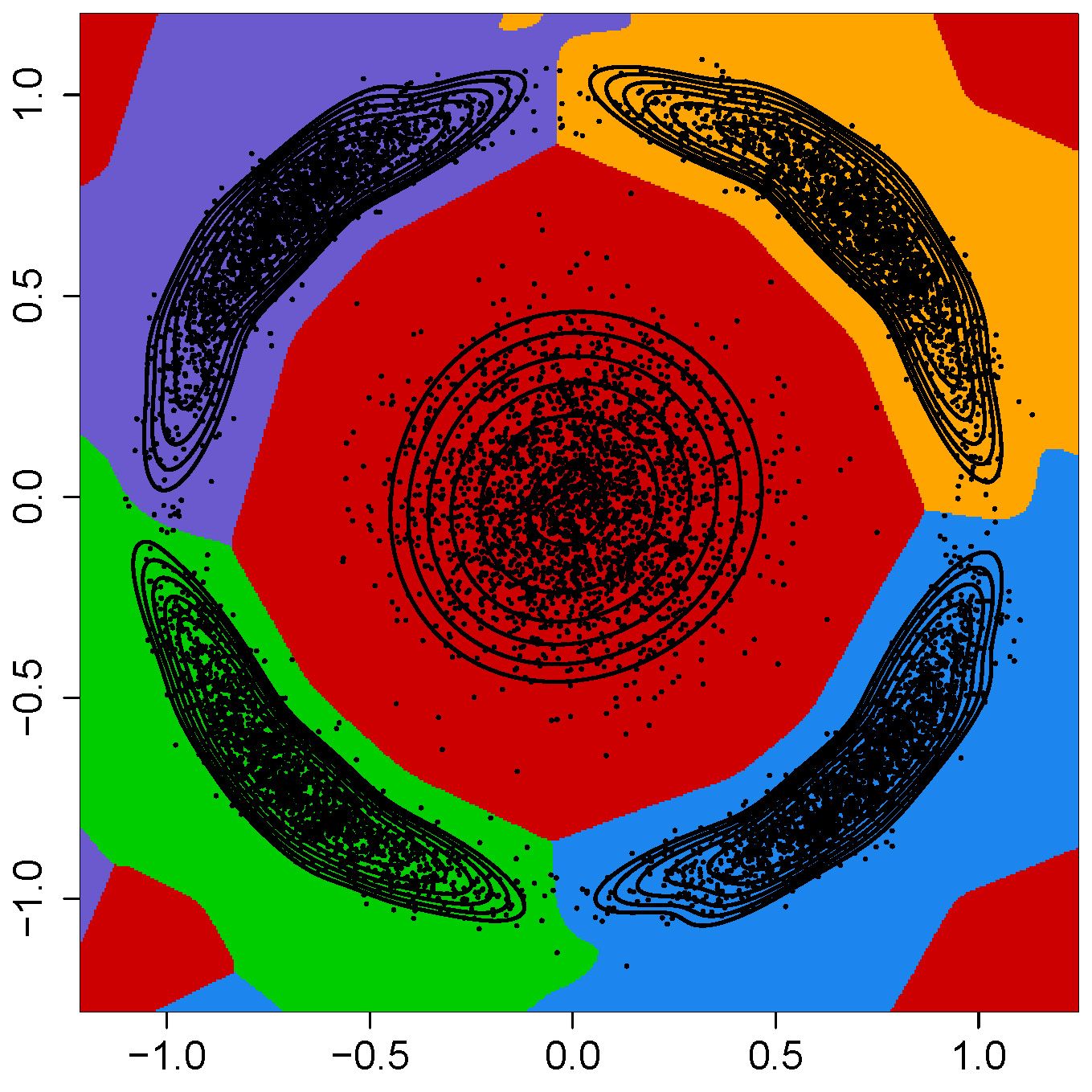}\\
\includegraphics[width=0.4\textwidth]{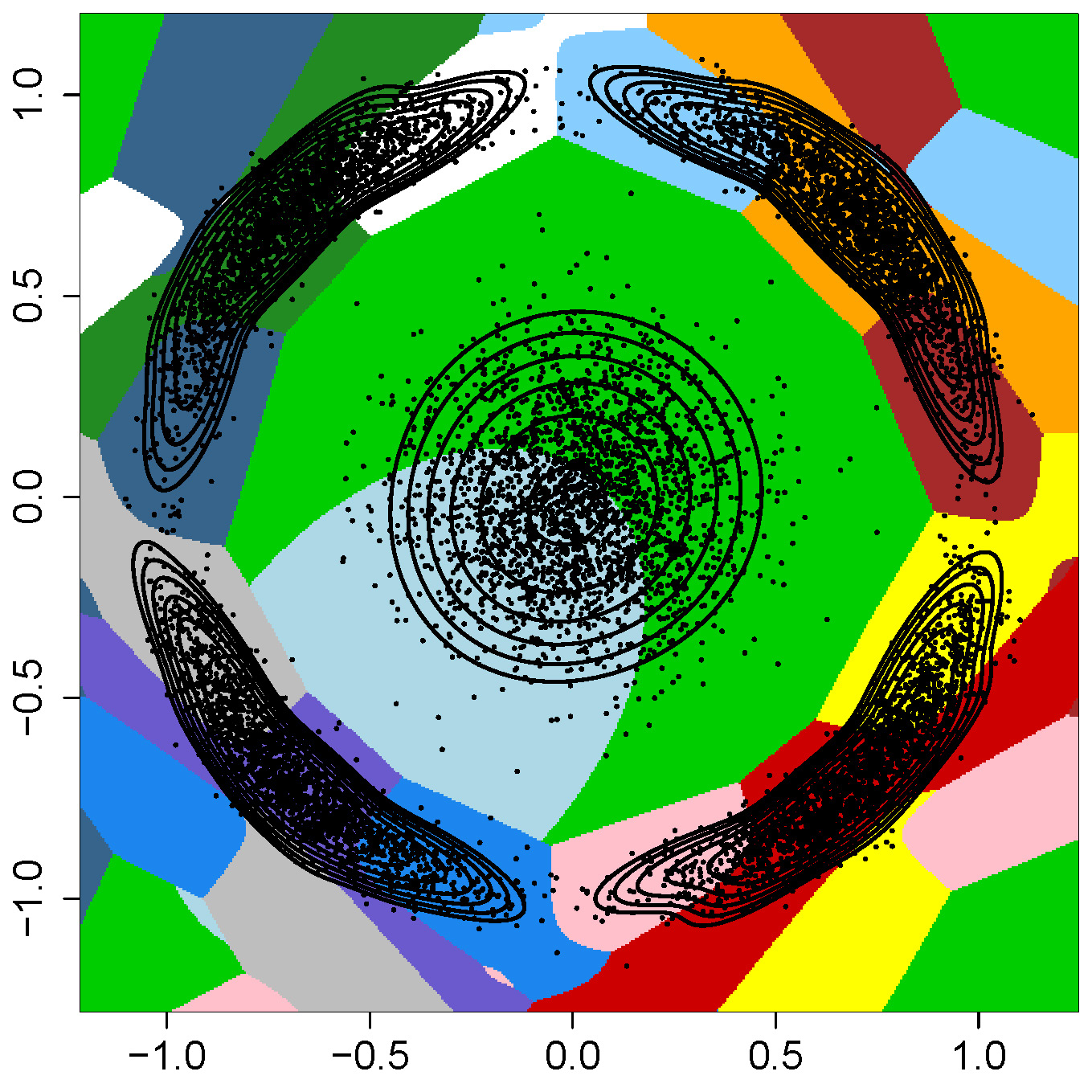}&\hspace{0.03\textwidth}\includegraphics[width=0.4\textwidth]{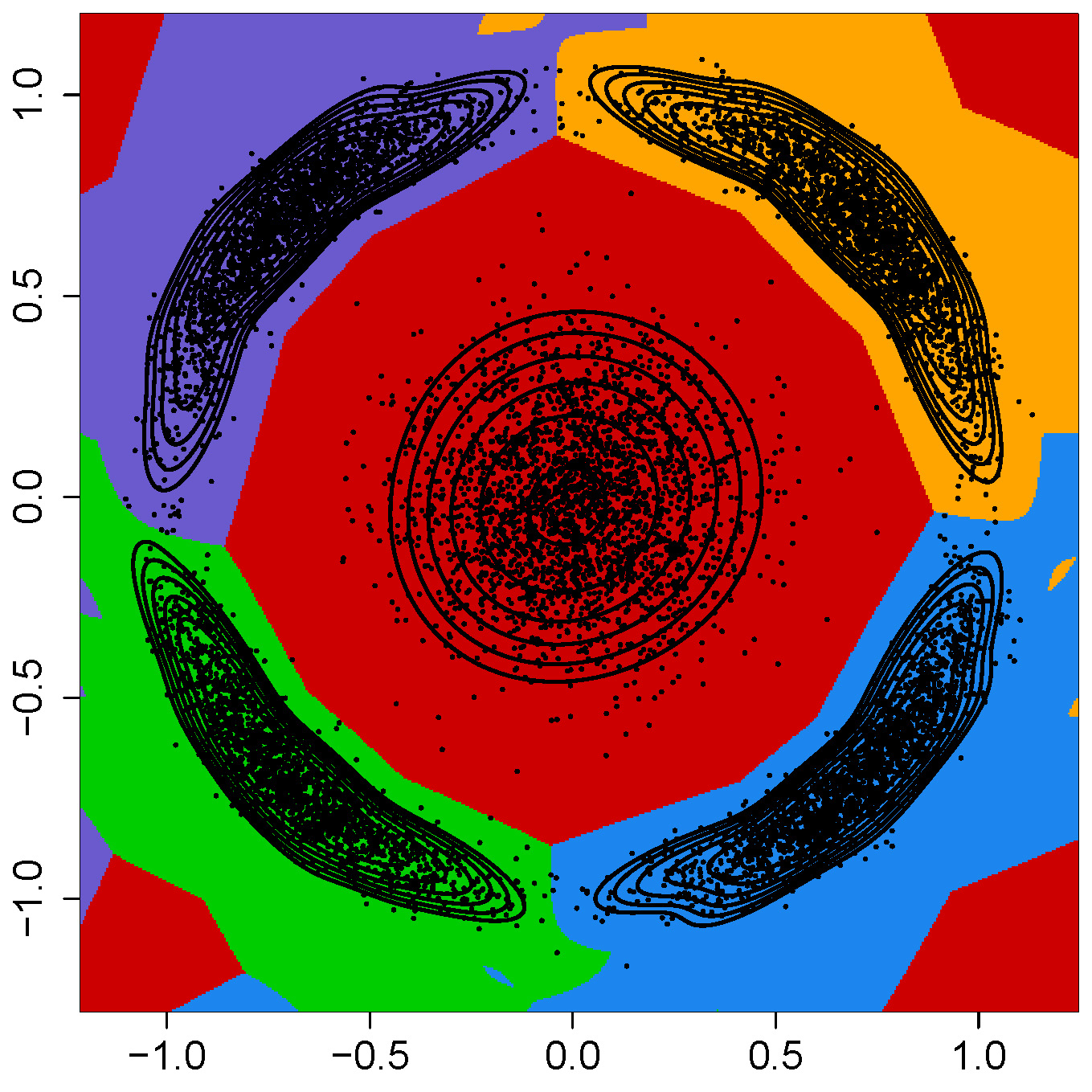}
\end{tabular}
\caption{Broken ring distribution. Top left, true modal clustering and true density contours. Top right, modal clustering on the normal mixture fit using Method 2. Bottom left, mixture model clustering with $G=9$ components. Bottom right, modal merging of the mixture components using Method 1.}
\label{fig:8}
\end{figure}

As in the previous example, the true modal clustering is shown in the top left plot of Figure \ref{fig:8}. A sample of size $n=5000$ was drawn from this distribution, and an application of the EM-BIC methodology resulted in a fit with $\widehat G=14$ normal components. Three normal components were needed to model each of the four crescent-shaped clusters, and two were used to fit the central one.

But despite $\widehat G=14$ normal components are needed to obtain a reasonably good mixture density estimate, the resulting density fit has only 5 modes. So the top right plot of Figure \ref{fig:8} shows the resulting modal cluster after applying Method 2 based on clustering the mixture density estimate from a modal point of view. The mosaic in the bottom left plot shows the clustering into 14 regions formed by assigning each point to its most likely component in the mixture density estimate, and finally the bottom right plot shows the result of merging the previous 14 regions into 5 modal clusters as a consequence of an application of Method 1. Again, the final modal clusterings that the two methods yield are very similar, with Method 1 being much faster (only the 14 component means were used as initial points for the mean shift algorithm) than Method 2, although the latter is somehow closer to the modal clustering philosophy.

\section{Discussion}

This paper illustrates how mixture modeling can be useful even if the final goal is to cluster the data according to a modal approach, instead of by mixture component assignment. Two different proposals are introduced for this task: one based on merging mixture components (Method 1) and a second one which uses the mean shift algorithm on the fitted normal mixture density to find the domains of attraction of the estimated density modes (Method 2).

The issue of the convergence of the mean shift algorithm is still not a fully solved problem. \citet{AG15} points out the incompleteness of several existing convergence proofs. In this paper, a new representation of the mean shift algorithm for non-isotropic normal mixture densities is provided, which allows to cast it as a quasi-Newton optimization method. This could be useful to address the convergence issue once again, since it could be tackled using the tools that are normally employed to study the convergence of such optimization methods \citep[see][Chapter 6]{DS96}.

As another open problem, even if \cite{LB00} showed that the mixture density estimator is consistent under mild assumptions, that does not guarantee that the number of modes of the mixture density estimator be a consistent estimator for the true number of density modes. Or, looking further afield, it would be even better if it could be proved that the modal clustering that is obtained from the mixture density estimate (using either Method 1 or Method 2) results in a consistent estimate of the true population modal clustering, in the sense indicated in \citet[][Section 4.3]{Ch15}.

Finally, all the examples included here try to illustrate the proposed methods from a qualitative point of view. A more detailed and exhaustive simulation study, in the spirit of \cite{CM14}, could throw some light on the evaluation of these methods from a more quantitative perspective.

\bigskip

\noindent{\bf Acknowledgments.} This paper was partly motivated by a question posed by Professor Ja\-cin\-to Mart\'\i n (Universidad de Extremadura). The author acknowledges the support of the Spanish Ministerio de Econom\'\i a y Competitividad grant MTM2013-44045-P and the Junta de Extremadura grant GR15013.

\bibliographystyle{apalike}

\begin{thebibliography}{}

\bibitem[Aliyari Ghassabeh, 2015]{AG15} Aliyari Ghassabeh, Y. (2015) A sufficient condition for the convergence of the mean shift
algorithm with Gaussian kernel. {\it Journal of Multivariate Analysis}, {\bf 135}, 1--10.

\bibitem[{Arias-Castro, Mason and Pelletier}, 2016]{AMP16} Arias-Castro, E., Mason, D. and Pelletier, B. (2016) On the estimation of the gradient lines of a density and the consistency of the mean-shift algorithm. {\it Journal of Machine Learning Research}, {\bf 17}, 1--28.

\bibitem[Azzalini and Bowman, 1990]{AB90} Azzalini, A. and Bowman, A.W. (1990) A look at some data on the Old Faithful geyser. {\it Applied Statistics}, {\bf 39}, 357--365.

\bibitem[Azzalini and Torelli, 2007]{AT07} Azzalini, A. and Torelli, N. (2007) Clustering via
    nonparametric density estimation. {\it Statistics and Computing}, {\bf 17}, 71--80.

\bibitem[Baudry, 2010]{B10} Baudry, J.-P. (2010) {\it S\'election de Mod\`{e}le pour la Classifcation Non Supervis\'ee. Choix du Nombre de Classes.} Ph D Thesis, Universit\'e Paris-Sud 11.

\bibitem[{Baudry {\it et al.}}, 2010]{Bal10} Baudry, J.-P., Raftery,  A.E., Celeux, G., Lo, K. and Gottardo, R. (2010) Combining mixture components
for clustering. {\it Journal of Computational and Graphical Statistics}, {\bf 19}, 332--353.

\bibitem[{Brinkman {\it et al.}}, 2007]{Bal07} Brinkman, R.R, Gasparetto, M., Lee, S.-J.J., Ribickas, A.J., Perkins, J., Janssen, W., Smiley, R. and
Smith, C. (2007) High-content flow cytometry and temporal data analysis for defining a cellular
signature of Graft-versus-Host Disease. {\it Biology of Blood and Marrow Transplantation}, {\bf 13}, 691--700.

\bibitem[Carlsson and M\'emoli, 2013]{CM13} Carlsson, G. and M\'emoli, F. (2013) Classifying clustering schemes. {\it Foundations of Computational Mathematics},
    {\bf 13}, 221--252.

\bibitem[Carreira-Perpi{\~n}{\'a}n, 2000]{CP00} Carreira-Perpi{\~n}{\'a}n, M.\'A. (2000) Mode-finding for mixtures of Gaussian distributions. {\it IEEE Transactions on Pattern Analysis and Machine Intelligence}, {\bf 22}, 1318--1323.

\bibitem[Carreira-Perpi{\~n}{\'a}n, 2006]{CP06} Carreira-Perpi{\~n}{\'a}n, M.\'A. (2006) Acceleration strategies for Gaussian mean-shift image segmentation. {\it IEEE Conference on Computer Vision and Pattern Recognition (CVPR 2006)}, 1160--1167.

\bibitem[Carreira-Perpi{\~n}{\'a}n, 2007]{CP07} Carreira-Perpi{\~n}{\'a}n, M.\'A. (2007) Gaussian mean shift is an EM algorithm. {\it IEEE Transactions on Pattern Analysis and Machine Intelligence}, {\bf 29}, 767--776.

\bibitem[Carreira-Perpi{\~n}{\'a}n and Williams, 2003a]{CPW03a} Carreira-Perpi{\~n}{\'a}n, M.\'A. and Williams, C.K.I. (2003a) On the number of modes of a Gaussian mixture. {\it Scale-Space Methods in Computer Vision}, pp. 625--640. Lecture Notes in Computer Science, vol. 2695, Springer-Verlag.

\bibitem[Carreira-Perpi{\~n}{\'a}n and Williams, 2003b]{CPW03b} Carreira-Perpi{\~n}{\'a}n, M.\'A. and Williams, C.K.I. (2003b) An isotropic Gaussian mixture can have more modes than components. Technical report EDI-INF-RR-0185, School of Informatics, University of Edinburgh, UK.

\bibitem[Chac\'on, 2012]{Ch12} Chac\'on, J.E. (2012) Identifying nonstandard group shapes in mixture model clustering through the mean shift algorithm. In {\it Programme and Abstracts of the 5th International Conference of the ERCIM Working Group on Computing \& Statistics}, 122.

\bibitem[Chac\'on, 2015]{Ch15} Chac\'on, J.E. (2015) A population background for
nonparametric density-based clustering. {\it Statistical Science}, {\bf 30}, 518--532.

\bibitem[Chac\'on and Duong, 2013]{CD13}
Chac\'on, J.E. and Duong, T. (2013)
Bandwidth selection for multivariate density derivative estimation, with applications to clustering and bump hunting.
{\it Electronic Journal of Statistics}, {\bf 7}, 499--532.

\bibitem[Chac\'on and Monfort, 2014]{CM14} Chac\'on, J.E. and Monfort, P. (2014) A comparison of bandwidth
selectors for mean shift clustering. In {\it Theoretical and Applied Issues in Statistics and Demography} (C. H. Skiadas, ed.), 47--59. International Society for the Advancement of Science and Technology (ISAST), Athens.

\bibitem[Comaniciu, 2003]{Co03} Comaniciu, D. (2003) An algorithm for data-driven bandwidth
    selection. {\it IEEE Transactions on Pattern Analysis and Machine Intelligence}, {\bf 25}, 281--288.

\bibitem[Comaniciu and Meer, 2002]{CM02} Comaniciu, D. and Meer, P. (2002) Mean shift: A robust
    approach toward feature space analysis. {\it IEEE Transactions on Pattern Analysis and Machine Intelligence}, {\bf 24},
    603--619.

\bibitem[{Cuevas, Febrero and Fraiman}, 2001]{CFF01} Cuevas, A., Febrero, M. and Fraiman, R. (2001)
    Cluster analysis: a further approach based on density estimation. {\it Computational Statistics and Data
    Analysis}, {\bf 36}, 441--459.

\bibitem[Dennis and Schnabel, 1996]{DS96} Dennis, J.E. and Schnabel, R.B. (1996) {\it Numerical Methods for Unconstrained Optimization and Nonlinear Equations}. SIAM, Philadelphia.

\bibitem[{Duong, Cowling, Koch and Wand}, 2008]{DCKW08} Duong, T., Cowling, A.,
    Koch, I. and Wand, M.P. (2008) Feature significance for multivariate
    kernel density estimation. {\it Computational Statistics and Data Analysis}, {\bf 52},
    4225--4242.

\bibitem[{Edelsbrunner, Fasy and Rote}, 2013]{EFR13} Edelsbrunner, H., Fasy, B.T. and Rote, G. (2013) Add isotropic Gaussian kernels at own risk: more and more resilient modes in higher dimensions. {\it Discrete \& Computational Geometry}, {\bf 49}, 797--822.

\bibitem[Edelsbrunner and Harer, 2008]{EH08} Edelsbrunner, H. and Harer, J. (2008) Persistent homology --- a survey. {\it Contemporary Mathematics}, {\bf 453}, 257--282.

\bibitem[Fraley and Raftery, 2002]{FR02} Fraley, C. and Raftery, A.E. (2002) Model-based clustering, discriminant analysis, and density estimation. {\it Journal of the American Statistical Association}, {\bf 97}, 611--631.

\bibitem[{Fraley, Raftery and Scrucca}, 2016]{FRS16} Fraley, C., Raftery, A.E. and Scrucca, L. (2016) {\it mclust: Gaussian Mixture Modelling for Model-Based Clustering, Classification, and Density Estimation}. R package version 5.2.


\bibitem[Fukunaga and Hostetler, 1975]{FH75} Fukunaga, K. and Hostetler, L.D. (1975) The estimation
    of the gradient of a density function, with applications in pattern recognition. {\it IEEE
    Transactions on Information Theory}, {\bf 21}, 32--40.

\bibitem[Hartigan, 1975]{H75} Hartigan, J.A. (1975) {\it Clustering Algorithms}. Wiley, New York.

\bibitem[Hennig, 2010]{H10} Hennig, C. (2010) Methods for merging Gaussian mixture components. {\it Advances in Data Analysis and Classification}, {\bf 4}, 3--34.

\bibitem[Li and Barron, 2000]{LB00} Li, J.Q. and Barron, A.R. (2000)  Mixture density estimation. In S.A. Solla, T.K. Leen and K-R. Mueller (Eds.), {\it Advances in Neural Information Processing Systems}, {\bf 12}, 279--285. MIT Press, Cambridge.

\bibitem[{Li, Hu and Wu}, 2007]{LHW07} Li, X., Hu, Z. and Wu, F. (2007) A note on the convergence of the mean shift. {\it Pattern recognition}, {\bf 40}, 1756--1762.

\bibitem[Lin, 2009]{L09} Lin, T.-I. (2009) Maximum likelihood estimation for multivariate skew normal mixture models. {\it Journal of Multivariate Analysis}, {\bf 100}, 257--265.

\bibitem[{Lin, Ho and Lee}, 2014]{LHL14} Lin, T.-I., Ho, H.J., Lee, C.-R. (2014) Flexible mixture modelling using the multivariate
    skew-$t$-normal distribution. {\it Statistics and Computing}, {\bf 24}, 531--546.

\bibitem[{Lo, Brinkman and Gottardo}, 2008]{LBG08}  Lo, K., Brinkman, R.R. and Gottardo, R. (2008) Automated gating of flow cytometry data via robust model-based clustering. {\it Cytometry A}, {\bf 73}, 321--332.

\bibitem[Priebe, 1994]{P94} Priebe, C.E. (1994) Adaptive mixtures. {\it Journal of the American Statistical Association}, {\bf 89}, 796--806.

\bibitem[Ray and Lindsay, 2005]{RL05} Ray, S. and Lindsay, B.G. (2005) The topography of multivariate normal mixtures. {\it Annals of Statistics}, {\bf 33}, 2042--2065.

\bibitem[Ray and Ren, 2012]{RR12} Ray, S. and Ren, D. (2012) On the upper bound of the number of modes of a multivariate normal mixture. {\it Journal of Multivariate Analysis}, {\bf 108}, 41--52.

\bibitem[{Rinaldo {\it et al.}}, 2012]{RSNW12} Rinaldo, A., Singh, A., Nugent, R. and Wasserman, L. (2012) Stability of density-based clustering. {\it Journal of Machine Learning Research}, {\bf 13}, 905--948.

\bibitem[Scrucca, 2016]{S16} Scrucca, L. (2016) Identifying connected components in Gaussian finite mixture models for clustering. {\it Computational Statistics and Data Analysis}, {\bf 93}, 5--17.

\bibitem[Stuetzle, 2003]{S03} Stuetzle, W. (2003) Estimating the cluster tree of a density by analyzing the minimal
spanning tree of a sample. {\it Journal of Classification}, {\bf 20}, 25--47.

\bibitem[Walther, 2003]{W03} Walther, G. (2003) Bikernel mixture analysis. In {\it Industrial Mathematics and Statistics} (J.C. Misra, ed.),  586--604. Narosa Publishing House, New Delhi.

\end{thebibliography}

\end{document}